\documentclass[preprint,review,12pt]{elsarticle}
\usepackage[utf8]{inputenc} 
\usepackage{hyperref}
\usepackage{longtable}
\usepackage{multirow}
\usepackage{lineno}
\usepackage{blindtext}
\usepackage{amssymb}
\usepackage{amsthm}
\usepackage{longtable}
\usepackage{subcaption}

\journal{Informatics in Medicine Unlocked}
\begin{document}

\title{
Risk markers by sex for in-hospital mortality in patients with acute coronary syndrome: a machine learning approach}                    

\author[1]{Blanca Vázquez\corref{cor1}}
\ead{blancavazquez2013@gmail.com}

\author[1]{Gibran Fuentes-Pineda}
\ead{gibranfp@unam.mx}

\author[1]{Fabian García}
\ead{fabian.garcia@iimas.unam.mx}

\author[2]{Gabriela Borrayo}
\ead{gabriela.borrayo@imss.gob.mx}

\author[3]{Juan Prohías}
\ead{prohiasmartinez@gmail.com}

\cortext[cor1]{Corresponding author}

\address[1]{Instituto de Investigaciones en Matemáticas Aplicadas y en Sistemas, Circuito Escolar s/n 4to piso, Ciudad Universitaria, Coyoacán, CDMX, 04510, Mexico.}

\address[2]{Programa ``A Todo Corazón", Centro Médico Nacional Siglo XXI, Instituto Mexicano del Seguro Social, Av. Cuauhtémoc 330, Doctores, Cuauhtémoc, CDMX, 06720, Mexico.}

\address[3]{Cardiocentro, Hospital Clínico Quirúrgico Hermanos Ameijeiras, Calle San Lázaro 701 esquina a Belascoaín, Padre Varela, La Habana, 10400, Cuba.}

\begin{abstract}
\textbf{Background:} 
Several studies have highlighted the importance of considering sex differences in the diagnosis and treatment of Acute Coronary Syndrome (ACS). However, the identification of sex-specific risk markers in ACS sub-populations has been scarcely studied. The present study aims to explore machine learning (ML) models to identify in-hospital mortality markers for women and men in ACS sub-populations collected from a public database of electronic health records (EHR). 

\noindent \textbf{Methods:} 
We extracted 1,299 patients with ST-elevation myocardial infarction (STEMI) and 2,820 patients with non-ST-elevation myocardial infarction (NSTEMI) from the Medical Information Mart for Intensive Care (MIMIC)-III database. We trained and validated mortality prediction models and used an interpretability technique to identify sex-specific markers for each sub-population.  

\noindent \textbf{Results:} 
The models based on eXtreme Gradient Boosting (XGBoost) achieved the highest performance: area under the curve (AUC) = 0.94 (95\% CI:0.84–0.96) for STEMI and AUC = 0.94 (95\% CI:0.80–0.90) for NSTEMI. For STEMI, the top markers in women are chronic kidney failure, high heart rate, and age over 70 years. For men, the top markers are acute kidney failure, high troponin T levels, and age over 75 years. However, for NSTEMI, the top markers in women are low troponin levels, high urea levels, and age over 80 years. For men, the top markers are high heart rate, creatinine levels, and age over 70 years. 

\noindent \textbf{Conclusions:} Our results show possible significant and coherent sex-specific risk markers of different ACS sub-populations by interpreting ML mortality models trained on EHRs. Differences are observed in the identified risk markers between women and men, highlighting the importance of considering sex-specific markers in implementing more appropriate treatment strategies and better clinical outcomes.
\end{abstract}


\maketitle
\newpage
\section{Introduction}
Acute Coronary Syndrome (ACS) is a leading cause of mortality and morbidity worldwide~\cite{mate_redondo_hospital_2019}. The two most common ACS conditions are ST-elevation myocardial infarction (STEMI) and non-ST-elevation myocardial infarction (NSTEMI). STEMI is a serious type of heart attack, caused by the complete blockage of one or more coronary arteries~\cite{foth_acute_2019}. In contrast, NSTEMI can be less serious because the blockage of the artery is partial. Although it can progress to STEMI if left untreated~\cite{mechanic_acute_2019,saleh_understanding_2018}.

ACS was long perceived as a health problem that predominantly affected men and, for this reason, women were frequently underrepresented in clinical trials. Consequently, significant differences in diagnostic criteria and treatment strategies between women and men have been widely reported in the last decade~\cite{linde_sex_2018,gao_gender_2019,leonarda_galiuto_gender_2017}. Several studies have highlighted the importance of sex-specific markers owing to differences in biological and physiological characteristics between, men and women~\cite{garcia_garcia_diferencias_2014, cheung_sex_based_2019, blom_women_2019}. This distinction could contribute to more appropriate treatment strategies, thus reducing risks and improving clinical outcomes~\cite{gao_gender_2019, borrayo_sanchez_stemi_2018}. According to~\cite{leonarda_galiuto_gender_2017}, the identification of risk markers between women and men represent a recent advance in the field of cardiovascular medicine, which must be studied in-depth.

At present, Electronic Health Records (EHR) provides opportunities to build evidence-based tools to support providers at the point of care~\cite{ranganath_deep_2016}. According to~\cite{hemingway_using_2017}, EHR assists providers to predict mortality accurately, anticipating major events, identifying risk factors, improving diagnosis, and improving patient outcomes. Generally, EHRs contain the patient’s medical history, such as demographic information, medication and allergies, laboratory test results, diagnoses, and so on. In the last years, Machine Learning (ML) methods have demonstrated promising results in accelerating markers identification and have become tools to leverage EHRs, analyze biomedical data, and facilitate clinical decisions in cardiovascular medicine~\cite{hemingway_using_2017}. For instance, Austin et al.~\cite{austin_regression_2012} trained ensemble-based methods using vital signs, physical examination data, and laboratory results to predict the probability of 30-day mortality in patients with ACS. The outcomes of their results show that age, systolic blood pressure, creatinine, and heart rate increase the risk of mortality. Similarly, Mcnamara et al.~\cite{mcnamara_predicting_2016} used logistic regression to predict in-hospital mortality in patients with myocardial infarction, and they identified age, systolic blood pressure, troponin, and heart failure as risk markers. Besides that, Chen et al.~\cite{chen_risk_2019} conducted a multivariate regression to predict in-hospital mortality rates in patients above 80 years. They found that the history of stroke, cardiac shock, Killip class III to IV, and elevated initial white blood cells were the markers of mortality. 

Although many studies use ML methods to identify risk markers for ACS, the distinction between women and men in ACS sub-populations has been scarcely explored. The present study aims to identify in-hospital mortality markers for women and men separately in STEMI and NSTEMI sub-populations using ML models. 

The major contributions of the paper are as follows.

\begin{itemize}
    \item We evaluated different ML models using EHRs data collected from the public database to predict mortality in patients with STEMI and NSTEMI. 
    \item We interpreted the mortality prediction models using a Shapley values-based technique to identify sex-specific markers in the ACS subpopulations.
    \item We validated the significance and coherence of the identified markers by applying a multivariable Cox regression, through expert cardiologists' assessments, and by comparing them with the analyzed markers reported in a previous longitudinal study.
\end{itemize}

The rest of the paper is structured as follows. Section 2 reviews the related works. Section 3 introduces details of the proposed methodology for training and evaluating the mortality models as well as for identifying the risk markers. Section 4 describes the baseline characteristics and reports the experimental results. The discussion of the experimental results is presented in Section 5. Finally, Section 6 concludes the paper by providing some remarks and limitations of the present work.

\section{Related works}
Risk markers identification for ACS has traditionally been conducted through retrospective and prospective studies. These studies developed scoring systems for the Thrombolysis in Myocardial Infarction (TIMI)~\cite{antman_timi_2000}, Platelet glycoprotein IIb/IIIa in Unstable angina: Receptor Suppression Using Integrilin Therapy (PURSUIT)~\cite{de_araujo_goncalves_timi_2005}, and Global Registry of Acute Coronary Events (GRACE)~\cite{fox_prediction_2006}. The TIMI risk score determines the likelihood of ischemic events and the risk of mortality in patients with NSTEMI and STEMI. Furthermore, the PURSUIT score predicts the risk of myocardial infarction or death 30 days after admission. Finally, GRACE estimates the risk of death in patients with ACS. In these systems, a set of risk factors are first established through clinical trials, and they are combined to obtain a score. However, some studies offer some disadvantages that limit the effectiveness of these scoring systems. For instance, the systems are usually calculated by hand using limited clinical features that are characterized by abnormal observations. Moreover, they rely on features that are not always readily available, and they do not distinguish markers based on patients’ sub-populations~\cite{harutyunyan_multitask_2019, barrett_building_2019}.

In the past decade, several studies have highlighted the importance of considering sex-specific markers in the care guidelines, risk factors, treatments, and pathophysiological mechanisms in ACS patients~\cite{cheung_sex_based_2019,the_eugenmed_gender_2016}. For instance, Wilkinson et al.~\cite{wilkinson_sex_2019} investigated the guidelines of care for STEMI and NSTEMI and their association with 30-day and 3-year mortality. They conclude that some sex-specific differences exist in the treatment strategies and women have a higher 30-day mortality risk than men. 

Lam et al.~\cite{lam_sex_2019} analyzed the sex differences in coronary heart diseases concerning by exploring the epidemiology, risk factors, pathophysiology, and response to therapy. They found that obesity, diabetes, and psychological stress are stronger risk factors in women than in men. Galiuto et al.~\cite{leonarda_galiuto_gender_2017} reported the symptoms and pathophysiological mechanisms underlying myocardial ischemia based on sex. They remarked that an increase in mortality in women was associated with a lack of appropriate management strategies. 

Rodriguez et al.~\cite{rodriguez_padial_differences_2021} studied in-hospital mortality and identified that women faced a higher risk of death after a STEMI and a lower risk of death after an NSTEMI. Some studies also investigate the sex differences in readmission rates and complications~\cite{cheung_sex_based_2019}, the opportunities to be resuscitated after a cardiac arrest~\cite{blom_women_2019}, and ACS mortality after a natural disaster~\cite{onose_sex_2017}. 

In recent years, ML algorithms have been successfully applied to identify risk markers in the clinical area. In particular, several studies have used ML models to identify markers for critical events in cardiovascular medicine~\cite{hemingway_using_2017}. Tokodi et al.~\cite{tokodi_sex_specific_2021} extracted sex-specific markers from patients undergoing cardiac resynchronization therapy using conditional inference random forest. For men, the identified markers were hemoglobin concentration, serum sodium, and serum creatinine. For women, the identified markers were age, serum sodium, and serum creatinine.

Similarly, Vinter et al.~\cite{vinter_role_2020} used logistic regression to identify markers for electrical cardioversion in patients with atrial fibrillation. For men, the most important markers were hemoglobin, age, and left atrial diameter. For women, age, hemoglobin, hypertension, and antiarrhythmic class III drugs are the most important markers.

A thorough search of the literature shows that a distinction of in-hospital mortality markers between women and men in ACS sub-populations using ML algorithms has been scarcely explored. 
In the present study, we aim to exploit the ML-based prediction models to identify sex-specific factors associated with a higher risk of in-hospital death in patients with STEMI and NSTEMI based on the information from an EHRs public database.

\section{Material and methods}
\begin{figure}[h!]
\centering
\includegraphics[scale=0.5]{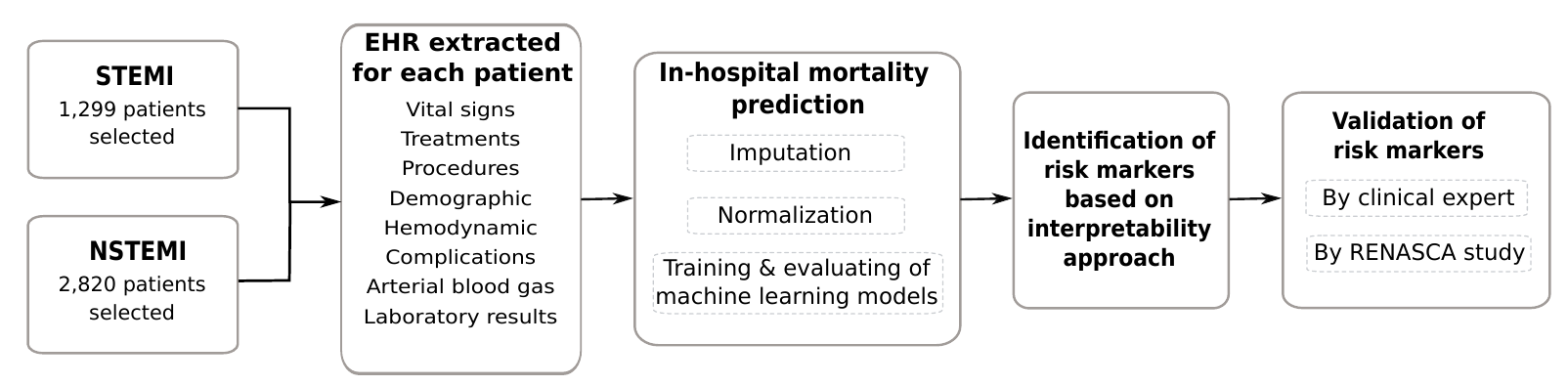}
\caption{Overall process for identifying risk markers in women and men for STEMI and NSTEMI patients using ML methods}
\label{fig:overall_process}
\end{figure}

We follow the conventional process using ML-based mortality prediction and risk marker identification in cardiovascular research~\cite{sengupta_proposed_2020}, as shown in Fig.~\ref{fig:overall_process}. For each ACS sub-population, we extracted a set of clinical features, trained, evaluated mortality prediction models, identified sex-specific risk markers using the prediction models with the highest performance, and validated the significance of the identified markers. Below we describe the detail of these steps.

\subsection{Study population}
In this study, we used the MIMIC-III database~\cite{johnson_mimic_iii_2016}, which is a publicly and freely available dataset that offers medical information about the Intensive Care Unit (ICU) for patients with diverse conditions of the Beth Israel Deaconess Medical Center between 2001 and 2012. From MIMIC-III, we extracted EHRs of patients admitted to ICU after suffering from a STEMI or NSTEMI. We used the codes 410.00-411.1 for selecting patients with STEMI and 410.70-410.72 for NSTEMI, as defined by the International Classification of Diseases, Ninth Revision (ICD-9). We describe the codes in detail in appendix A and B.

\subsection{Data extraction}
We extracted eight groups of clinical features, namely demographic characteristics, laboratory results, vital signs, arterial blood gas, hemodynamics, complications, treatments, and procedures. We used one-hot encoding for categorical features, and continuous features were represented with the minimum, maximum, and average values, as commonly used for predicting mortality (e.g.~\cite{barrett_building_2019}). However, some features were not found in the MIMIC-III database, such as precordial leads, to compare STEMI with NSTEMI. Consequently, we extracted a total of 191 features for STEMI and 201 for NSTEMI. Table~\ref{tab:clinical_variables} describes the extracted features of the above-mentioned groups in detail.

We filled missing values with the mean of the observed values of the corresponding feature. We paid special attention to features associated with myocardial infarction size and heart injuries (e.g., troponins, pulmonary artery pressure, and leads)~\cite{hemingway_using_2017}. The values of these features were gathered from the clinical notes using a set of regular expressions. Regarding the variability of the range of values between features, we performed feature-wise normalization on each sample by subtracting the mean and dividing it by the standard deviation. Finally, the results of the data set were split randomly into non-overlapping training and test sets, consisting of 80\% and 20\%, respectively.

\begin{center}
\small{
\begin{longtable}{ll}
\caption{\label{tab:clinical_variables} Extracted clinical features during the first 24 hours at admission for each ACS sub-population}
\\ \hline
\textbf{Clinical set} 
& 
\textbf{Features}
\\ \hline
Demographic
& \begin{tabular}[c]{@{}l@{}}
gender, age, admission type (elective, emergency, urgent), \\status (divorced, married, single, widow), weight admit
\end{tabular}
\\ \hline
Vital signs
& \begin{tabular}[c]{@{}l@{}}
heart rate, blood pressure (systolic, diastolic, mean), \\respiratory rate, oxygen saturation, temperature
\end{tabular}
\\ \hline
Laboratory results
& \begin{tabular}[c]{@{}l@{}}
troponin T, troponin I, anion gap, albumin, bands, urea, \\uric acid, creatinine, fibrinogen, sodium, triglycerides, \\glucose, white blood cells, partial thromboplastin time, \\neutrophils, lymphocytes, basophils, monocytes, protein \\creatinine ratio, eosinophils, international normalized \\ratio, prothrombin time, platelets, potassium, positive \\end-expiratory pressure, cholesterol (total, hdl, ldl), \\hemoglobin a1c, hematocrit, hemoglobin, c-reactive, \\creatine kinase ck, creatine kinase MB

\end{tabular}
\\ \hline
Hemodynamic
& \begin{tabular}[c]{@{}l@{}}
cardiac out, intracranial pressure, devices beat rate \\(left, right), pulmonary artery pressure (systolic, diastolic, \\mean), central venous pressure, ventricular assist device \\(left, right), pulmonary capillary wedge pressure, mixed \\venous oxygen saturation, pulmonary artery line, \\ventricular assist
\\ \end{tabular}
\\ \hline
Arterial blood gas
& \begin{tabular}[c]{@{}l@{}}
alveolar-arterial gradient, base excess, SO2, PO2, \\PCO2, Total CO2, chloride, calcium, lactate, FiO2, \\bicarbonate, PH
\end{tabular}
\\ \hline
Treatments
& \begin{tabular}[c]{@{}l@{}}
aspirin, clopidogrel bisulfate, enoxaparin, heparin, oral \\nitrates statins, fibrates, beta-blockers, amiodarone, ace \\inhibitors, Angiotensin II receptor blockers, insulin, \\diuretics, calcium antagonist, potassium chloride, \\oral glucose low drugs, digoxin, dobutamine, \\dopamine, warfarin, vancomycin
\end{tabular}
\\ \hline
Procedures
& \begin{tabular}[c]{@{}l@{}}
coronary arteriography using two catheters, injection or \\infusion of platelet inhibitor, combined right and left \\heart cardiac catheterization, circulation auxiliary to \\open-heart surgery, replacement of tracheostomy tube, \\angiocardiography of left heart structures, insertion \\of the endotracheal tube, angiocardiography \\of right heart structures, the extracorporeal implant of \\pulsation balloon, venous catheterization, coronary \\arteriography using a single catheter, arterial \\catheterization, insertion of the temporary transvenous \\pacemaker system
\end{tabular}
\\ \hline
Complications
& \begin{tabular}[c]{@{}l@{}}
ventricular fibrillation, ventricular tachycardia, atrial \\fibrillation, atrioventricular block, angina, left bundle \\branch block, right bundle branch block, cardiogenic \\shock, pericarditis, renal failure, hypertension, \\mitral regurgitation, cardiac arrest, diabetes, congestive \\heart failure, chronic airway obstruction, aneurysm, \\cerebrovascular accident, leads
(i, ii, iii, v1, v2, \\v3, v4, v5, v6, avf, avr, avl, f). 
For STEMI: leads \\(v1r, v2r), qtc wave. For NSTEMI: leads (lv, l, v), \\septal rupture, anterolateral, lateral, precordial, \\inferolateral, anterior, mid-lateral, posterolateral, \\inferior, hypertrophy, left ventricular, waves (r, qt, \\inverted t, qrs, rv).
\end{tabular}
\\ \hline
\end{longtable}
}
\end{center}

\subsection{In-hospital mortality prediction models}
We trained and evaluated data using linear and nonlinear ML algorithms to predict mortality~\cite{tokodi_sex_specific_2021,liu_predicting_2021}. ML algorithms used for prediction were Logistic Regression (LR), Support Vector Machines (SVM), Random Forest (RF), and eXtreme Gradient Boosting (XGB). 

For LR, we used the $saga$ optimizer, $\ell_1$, $\ell_2$, and $elasticnet$ norms for weight penalization. For SVM, we explored the $\ell_1$ and $\ell_2$ weight penalization norms. For both SVM and LR, we explore different strengths $C$ of penalization on a logarithmic scale in a range from $-3$ to $3$. For RF, the base-2 logarithm of the available features was used as the maximum number of features for each split, and the quality of the split was measured with the $gini$ function. For XGB, the $\ell_1$ and $\ell_2$ norms for weight penalization and 0.05, 0.1, and 0.5 for the learning rate were considered. We evaluated 0.3, 0.4, 0.8, and 0.9 as subsample ratios to randomly sample the training data prior to growing the decision trees. In addition, we examined 0.3 and 0.5 for dropout rate and values 10–50 for $\gamma$. Models with 50, 100, and 200 trees with a maximum depth of 2, 4, and 6 nodes were evaluated for both RF and XGB. We used weighted loss functions for all methods to mitigate the class imbalance problem. Specifically, the loss for the class c is weighted by

\begin{equation}\label{eq:cweight}
   w_{c} = \frac{n}
   {2 \cdot n_{c}}
\end{equation}

\noindent where $w_{c}$ is the weight of the class $c\in \{0,1\}$; $n$ is the total number of samples in the dataset, and $n_{c}$ is the number of samples for class $c$.

For model selection, we relied on grid search to compare the prediction performance based on the 10 repetitions of stratified 10-fold cross-validation on the training set. We computed the Area Under (AUC) the Receiver Operating Characteristic (ROC) curve and select the model with the highest mean of cross-validated AUC for each subpopulation. Finally, we estimate the prediction performance of these models over the test set.

We evaluated models trained with all clinical features (Table~\ref{tab:clinical_variables}) and models trained with different feature groups separately, to study the impact of each group on the mortality prediction. We compared the performance of the prediction models with the mean of the highest cross-validated AUC against the GRACE score, the most common clinical score to predict mortality for ACS. We extracted the values of all the GRACE markers, obtained the score of each patient using the GRACE scale~\cite{granger_predictors_2003}, and computed the ROC curve and AUC from all the scores. The source code for all the reported experiments is available at~\url{https://github.com/blancavazquez/Riskmarkers_ACS}. 

\subsection{Identification of risk markers by sex for STEMI and NSTEMI patients} 
We adopt an interpretability approach to identify risk markers. In particular, we apply the SHapley Additive exPlanations (SHAP) algorithm~\cite{lundberg_unified_2017} to interpret the output of the prediction models. The SHAP algorithm has been recently exploited to identify markers of chronic kidney disease~\cite{lundberg_local_2020}  and hypoxemia risk~\cite{lundberg_explainable_2018}.

The goal of SHAP is to explain the prediction of an instance $x$ by computing the contribution of each feature to provide a prediction. To achieve this objective, SHAP computes the Shapley Values using the coalitional game theory~\cite{roth_shapley_1988}, where games have competing teams composed of $p$ players each. Since each player contributes differently to win a game, the payout is distributed fairly among all the players. Specifically, the Shapely value $\phi_{j}(val)$ is the fair payout that a player $j$ receives for a game and is defined as:

\begin{equation}\label{ref:shapley_values}
\phi_{j}(val)= \sum_{S\subseteq\{{1},\ldots,{p}\}\backslash \{{j}\}} 
\frac{|S|! (p-|S|-1)!}
{p!}
(val(S\cup \{{j}\}) - val(S))
\end{equation}

\noindent where the summation is based on all possible subsets $S$ of the remaining players; $val$ is a function that returns the contribution of a given subset, and $p$ is the total number of players.

From the SHAP algorithm, it is possible to compute the importance of feature by averaging the per-feature absolute Shapley values over all the datasets are as follows: 

\begin{equation}\label{ref:feature_importance}
   I_{j}= \sum_{i=1}^{n} |\phi_{j}^{(i)}| 
\end{equation}

\noindent where $n$ is the number of instances in the dataset. Hence, the features with large absolute Shapley values are important for the model’s predictions.

\subsection{Validation of risk markers}
A set of cardiologists evaluated the reliability and relevance of the identified markers, to determine whether the markers are useful for predicting mortality in routine clinical practice. Additionally, we compared the identified markers with a longitudinal-cohort study focusing on STEMI and NSTEMI for a real-world study in Mexico called RENASCA~\cite{borrayo_sanchez_stemi_2018}.

\subsection{Statistical analysis}
Continuous features were compared with the Student’s T-test and categorical features were compared using the Chi-square test. We used the Delong test to compare the ROC curves of ML models and the GRACE score. We used the McNemar test to compare errors rates between ML models and the GRACE score on the test sets. When the errors are different, the test suggests a statistically significant difference between the ML model and GRACE (p \textless~0.05). We conduct a survival analysis with multivariate Cox regression to identify the features that have a statistically significant association with mortality.

\section{Results}
\subsection{Baseline Characteristics}
Our cohorts consist of 1,299 patients diagnosed with STEMI and 2,820 with NSTEMI, with a length of stay of more than 24 hours. Overall, for STEMI, 65\% of the patients were men and 35\% were women, with an average age of 67.26. Contrarily,  for NSTEMI, 58\% were men, and 42\% were women, and the average age was 72.29. The mortality rate for STEMI was 6.77\% and 9.21\% for NSTEMI. However, atrial fibrillation and diuretics are the most common complications and treatments, respectively recorded in both populations. Table~\ref{tab:statistics_patients} summarizes the baseline characteristics.

\begin{table}[h!]
\caption{\label{tab:statistics_patients}Baseline characteristics for patients with STEMI and NSTEMI}
\begin{center}
\scalebox{0.51}{
\begin{tabular}{lcccccccc}
\hline
\multicolumn{1}{c}{\multirow{2}{*}{Feature}}
& \multicolumn{4}{c}{STEMI}
& \multicolumn{4}{c}{NSTEMI}                               
\\ \cline{2-9} 
\multicolumn{1}{l}{}
& \multicolumn{1}{c}{\begin{tabular}[c]{@{}c@{}}Total cohort\\ n = 1299\end{tabular}} & \multicolumn{1}{c}{\begin{tabular}[c]{@{}c@{}}Women\\ n = 460 (35\%)\end{tabular}} & \multicolumn{1}{c}{\begin{tabular}[c]{@{}c@{}}Men\\ n = 839 (65\%)\end{tabular}} & \multicolumn{1}{l}{P-value} & \multicolumn{1}{c}{\begin{tabular}[c]{@{}c@{}}Total cohort\\ n = 2820\end{tabular}} & \multicolumn{1}{c}{\begin{tabular}[c]{@{}c@{}}Women\\ n = 1176 (42\%)\end{tabular}} & \multicolumn{1}{c}{\begin{tabular}[c]{@{}c@{}}Men\\ n = 1644 (58\%)\end{tabular}} 
& \multicolumn{1}{c}{P-value} \\ \hline
Age
& 67.26 ± 13.86
& 72.89 ± 13.33
& 64.17 ± 13.16
& \textless{} 0.001
& 72.29 ± 13.38
& 74.40 ± 12.50
& 70.66 ± 12.60
& \textless{} 0.001
\\
Weight (kg)
& 79.65 ± 17.83
& 72.02 ± 15.83
& 86.19 ± 16.58
& 0.325
& 81.18 ± 17.67
& 72.92 ± 16.90
& 84.46 ± 16.89
& 0.019
\\
\multicolumn{9}{l}{\textbf{Risk factors}}
\\
Hypertension
& 647 (49.80\%)
& 228 (49.56\%)
& 419 (49.94\%)
& 0.284
& 1,313 (46.56\%)
& 585 (49.74\%)
& 728 (44.28\%)
& 0.3
\\
Diabetes
& 292 (22.47\%)
& 103 (22.39\%)
& 189 (22.52\%)
& 0.097
& 800 (28.36\%)
& 331 (28.14\%)
& 469 (28.52\%)
& 0.015
\\
Smoking
& 170 (13.08\%)
& 43 (9.34\%)
& 127 (15.13\%)
& 0.006
& 198 (7.02\%)
& 69 (5.86\%)
& 129 (7.84\%)
& \textless{} 0.001
\\
\multicolumn{9}{l}{\textbf{Hemodynamic assessment}}
\\
Heart rate (bpm)
& 80.81 ± 14.33
& 82.08 ± 14.38
& 80.11 ± 14.25
& \textless{} 0.001
& 83.72 ± 14.09
& 84.13 ± 14.47
& 83.42 ± 13.79
& \textless{} 0.001
\\
Respiration rate (bpm)
& 19.57 ± 8.98
& 20.56 ± 11.23
& 19.17 ± 7.84
& 0.1
& 19.06 ± 3.89
& 19.26 ± 4.06
& 18.91 ± 3.75
& \textless{} 0.001
\\
Sysbp (mmHg)
& 112.09 ± 14.35
& 112.21 ± 14.38
& 112.27 ± 13.92
& \textless{} 0.001
& 116.25 ± 15.53
& 117.37 ± 16.48
& 115.44 ± 14.75
& \textless{} 0.001
\\
Diasbp (mmHg)
& 61.00 ± 9.64
& 57.20 ± 8.72
& 63.06 ± 9.50
& 0.004
& 57.62 ± 11.02
& 56.02 ± 9.93
& 58.77 ± 11.59
& 0.009
\\
\multicolumn{9}{l}{\textbf{Biochemistry determinations}}                  
\\
HbA1c (g/dl)
& 6.56 ± 0.76
& 6.55 ± 0.58
& 6.56 ± 0.83
& 0.778
& 6.73 ± 0.52
& 6.72 ± 0.49
& 6.72 ± 0.52
& \textless{} 0.001
\\
Creatinine (µmol/L)
& 3.23 ± 10.94
& 3.32 ± 9.48
& 3.17 ± 11.66
& \textless{} 0.001
& 5.58 ± 12.77
& 4.80 ± 10.38
& 6.14 ± 14.21
& \textless{} 0.001
\\
CK-MB  (U/L)
& 197.37 ± 212.11
& 184.70 ± 195.37
& 204.39 ± 220.5
& \textless{} 0.001
& 52.57 ± 71.20
& 47.39 ± 59.98
& 56.28 ± 78.03
& \textless{} 0.001
\\
Troponin T
& 8.31 ± 12.05
& 8.19 ± 9.18
& 8.37 ± 13.38
& 0.08
& 3.04 ± 7.10
& 3.07 ± 8.82 
& 3.01 ± 5.54
& 0.012
\\
\multicolumn{9}{l}{\textbf{Complications}}
\\
Atrial fibrillation
& 323 (24.86\%)
& 129 (28.04\%)
& 194 (23.12\%)
& \textless{} 0.001
& 962 (34.11\%)
& 397 (33.75\%)
& 565 (34.36\%)
& 0.08
\\
Acute renal failure
& 159 (12.24\%)
& 64 (13.91\%)
& 95 (11.32\%)
& \textless{} 0.001
& 760 (26.95\%)
& 317 (26.95\%)
& 443 (26.94\%)
& \textless{} 0.001
\\
RBBB
& 100 (7.69\%)
& 42 (9.13\%)
& 58 (6.91\%)
& 0.867
& 256 (9.70\%)
& 100 (8.50\%)
& 156 (9.48\%)
& 0.038
\\
LBBB
& 60 (4.6\%)                                                     
& 18 (3.91\%) 
& 42 (5.0\%)
& \textless{} 0.001
& 254 (9.0\%)
& 105 (8.92\%)
& 149 (9.06\%)
& 0.076
\\
\multicolumn{9}{l}{\textbf{Treatments}}
\\
ACE inhibitors
& 360 (27.71\%)
& 103 (22.39\%)
& 257 (30.63\%)
& \textless{} 0.001
& 329 (11.66\%)
& 134 (11.39\%)
& 195 (11.86\%)
& \textless{} 0.001
\\
Diuretics
& 341 (15.50\%)
& 130 (28.26\%)
& 211 (25.14\%)
& 0.418
& 1094 (38.78\%)
& 456 (38.77\%)
& 638 (38.87\%)
& 0.003
\\
Aspirin
& 224 (17.24\%)
& 79 (17.17\%)
& 145 (17.28\%)
& 0.05
& 744 (26.38\%)
& 292 (24.82\%)
& 452 (27.49\%)
& 0.116
\\
Clopidogrel
& 152 (11.70\%)
& 50 (10.86\%)
& 102 (12.15\%)
& 0.362
& 286 (10.14\%)
& 112 (9.52\%)
& 174 (10.58\%)
& \textless{} 0.001
\\
\textbf{Average stay (days)}
& 4.39
& 4.57 
& 4.30
&
& 5.12
& 5.26
& 5.02
&
\\
\begin{tabular}[c]{@{}l@{}}\textbf{Number of patients} \\ \textbf{expired (first 24 hours)}\end{tabular}
& 88 (6.77\%)
& 42 (9.13\%)
& 46 (5.48\%)
& 
& 260 (9.21\%)
& 126 (10.71\%)
& 134 (8.15\%)
& 
\\ \hline
\end{tabular}
}
\end{center}
{\raggedright 
\tiny
ACE inhibitors: Angiotensin-converting enzyme inhibitors; bpm: beats per minute; bpm: breaths per minute; CK-MB: Creatine kinase MB fraction; Diasbp: Diastolic blood pressure;g/dl: grams per deciliter; HbA1c: glycated hemoglobin; LBBB: Left bundle branch block; mmHg: millimeters of mercury; RBBB: Right bundle branch block; Sysbp: Systolic blood pressure; U/L: unit per liter; mol/L: micromol per liter. Data shown are mean – standard deviation for continuous features and as a percentage for categorical features
\par}
\end{table}

\subsection{Performance of in-hospital mortality prediction models}
In Table~\ref{tab:auc_models}, we present the mean of cross-validated AUC for the LR, RF, SVM, and XGB models in each feature group. Generally, the XGB models obtained the highest mean AUC, except for demographic, treatments, and complications. In addition, models trained with all features (combined) achieved a higher AUC than models trained with a single group. For both STEMI and NSTEMI, we selected the XGB models trained with all the extracted features. For STEMI, the hyperparameters of the selected model were as follows: maximum depth = 4 and $\ell_2$ regularization rate = 0.6. Contrarily for NSTEMI, the hyperparameters of the selected model were maximum depth = 6 and $\ell_2$ regularization rate = 0.2. For both models, the minimum loss reduction was 10; the learning rate was 0.1; the drop rate was 0.5; the number of trees was 250; the subsample ratio was 0.9, and the $\ell_1$ regularization rate was 0.5. 

\begin{center}
\small{
\begin{longtable}{llll}
\caption{\label{tab:auc_models} Performance of LR, RF, SVM, and XGB with the selected STEMI and NSTEMI hyperparameters using different clinical sets.}
\\ \hline
\textbf{\textbf{\begin{tabular}[c]{@{}c@{}}
Clinical set\end{tabular}}} 
& \textbf{Model}
& \textbf{\textbf{\begin{tabular}[c]{@{}c@{}}
STEMI\\ AUC ± STD \end{tabular}}} 
& \textbf{\textbf{\begin{tabular}[c]{@{}c@{}}
NSTEMI\\ AUC ± STD \end{tabular}}} 
\\ \hline
\multirow{4}{*}{\begin{tabular}[c]{@{}l@{}}Demographic\end{tabular}}
& LR
& 0.65 ± 0.09
& 0.64 ± 0.07
\\
& RF
& 0.57 ± 0.08
& 0.66 ± 0.06
\\
& SVM
& \textbf{0.66 ± 0.09}
& 0.67 ± 0.05
\\
& XGB
& 0.62 ± 0.09
& \textbf{0.69 ± 0.05}
\\ \hline
\multirow{4}{*}{\begin{tabular}[c]{@{}l@{}}Vital signs\end{tabular}}
& LR
& \textbf{0.88 ± 0.05}
& 0.91 ± 0.02
\\
& RF
& 0.81 ± 0.07
& 0.89 ± 0.03
\\
& SVM
& 0.87 ± 0.05
& 0.86 ± 0.03
\\
& XGB
& \textbf{0.88± 0.05}
& \textbf{0.92 ± 0.02}
\\ \hline
\multirow{4}{*}{\begin{tabular}[c]{@{}l@{}}Laboratory results\end{tabular}}
& LR
& 0.83 ± 0.06
& 0.74 ± 0.06
\\ 
& RF
& 0.80 ± 0.08
& 0.70 ± 0.05
\\ 
& SVM
& 0.79 ± 0.09
& 0.73 ± 0.06
\\ 
& XGB
& \textbf{0.87 ± 0.06}
& \textbf{0.76 ± 0.02}
\\ \hline
\multirow{4}{*}{\begin{tabular}[c]{@{}l@{}}Hemodynamic\end{tabular}}
& LR
& \textbf{0.61 ± 0.09}
& 0.66 ± 0.06
\\
& RF
& 0.53 ± 0.10
& 0.64 ± 0.06
\\ 
& SVM
& 0.59 ± 0.11
& 0.66 ± 0.06
\\ 
& XGB
& \textbf{0.61 ± 0.09}
& \textbf{0.70 ± 0.05}
\\ \hline
\multirow{4}{*}{\begin{tabular}[c]{@{}l@{}}Arterial blood gas\end{tabular}}
& LR
& 0.70 ± 0.13
& 0.69 ± 0.06
\\ 
& RF
& 0.71 ± 0.11
& 0.68 ± 0.06
\\ 
& SVM
& 0.72 ± 0.10
& 0.69 ± 0.06
\\ 
& XGB
& \textbf{0.81 ± 0.08}
& \textbf{0.72 ± 0.06}
\\ \hline
\multirow{4}{*}{\begin{tabular}[c]{@{}l@{}}Treatments\end{tabular}}
& LR
& \textbf{0.69 ± 0.08}
& 0.65 ± 0.06
\\ 
& RF
& 0.65 ± 0.09
& 0.63 ± 0.06
\\ 
& SVM
& 0.66 ± 0.10
& 0.65 ± 0.06
\\ 
& XGB
& 0.68 ± 0.08
& 0.65 ± 0.06
\\ \hline
\multirow{4}{*}{\begin{tabular}[c]{@{}l@{}}Procedures \end{tabular}}
& LR
& 0.75 ± 0.10
& 0.69 ± 0.07
\\ 
& RF
& 0.79 ± 0.08
& 0.74 ± 0.04
\\ 
& SVM
& 0.80 ± 0.07
& 0.76 ± 0.04
\\
& XGB
& \textbf{0.82 ± 0.06}
& \textbf{0.77 ± 0.04}
\\ \hline
\multirow{4}{*}{\begin{tabular}[c]{@{}l@{}}Complications\end{tabular}}
& LR
& \textbf{0.83 ± 0.06}
& 0.73 ± 0.05
\\ 
& RF
& 0.72 ± 0.11
& 0.68 ± 0.07
\\ 
& SVM 
& 0.74 ± 0.10
& 0.71 ± 0.06
\\ 
& XGB
& 0.82 ± 0.06
& \textbf{0.74 ± 0.05}
\\ \hline
\multirow{4}{*}{\begin{tabular}[c]{@{}l@{}}Combined\end{tabular}}
& LR
& 0.91 ± 0.04
& 0.92 ± 0.03
\\ & RF
& 0.88 ± 0.04
& 0.89 ± 0.02
\\ & SVM
& 0.88 ± 0.05
& 0.91 ± 0.03
\\ & XGB
& \textbf{0.95 + 0.03}
& \textbf{0.94 ± 0.02}
\\ \hline
\end{longtable}
{\raggedright 
\tiny{LR, Logistic Regression; 
SVM, Support Vector Machines; 
XGB, eXtreme Gradient Boosting; 
RF, Random Forest; STD, standard deviation; `combined'  means to join all the clinical features extracted to train the model.
\par}}}
\end{center}

\begin{figure*}
\centering
\subcaptionbox{STEMI}{\includegraphics[scale=0.38]{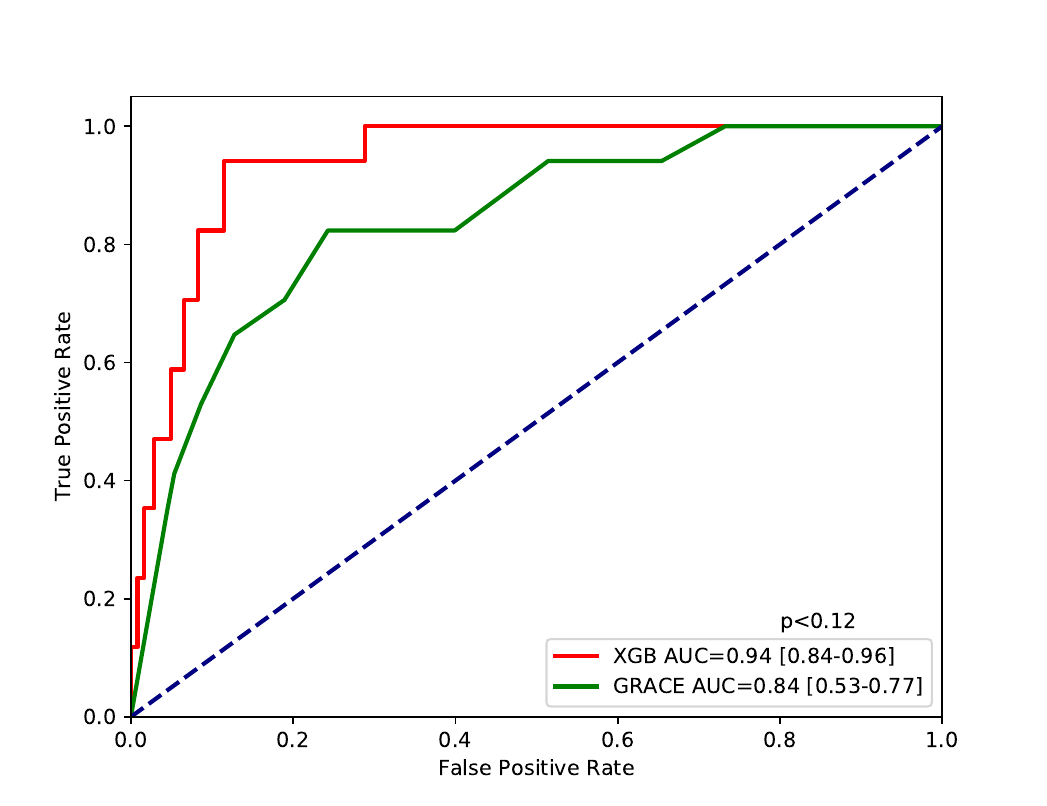}}
\subcaptionbox{NSTEMI}{\includegraphics[scale=0.38]{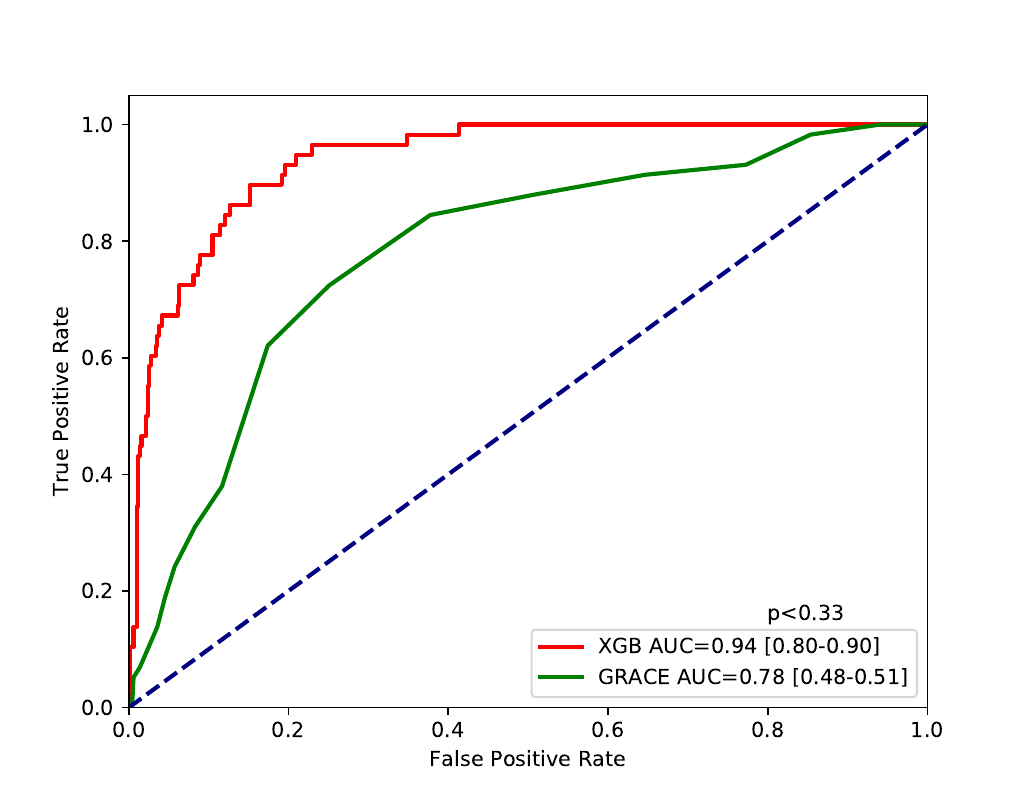}}
\caption{ROC curves for predicting in-hospital mortality within the first 24 hours of admission in the ICU for a STEMI (a) or NSTEMI (b). XGB-based models achieved higher AUCs than the GRACE score.}
\label{fig:roc}
\end{figure*}

Finally, we computed the ROC curve and corresponding AUC for the selected XGB model and the GRACE score in the test set (Figure~\ref{fig:roc}). As observed, XGB models achieved a significantly higher score for the test AUC than the GRACE score. For STEMI, the test AUC of the selected XGB model was 0.94 (95\% CI:0.84-0.96), while GRACE achieved 0.84 (95\% CI:0.53-0.77). In contrast, for NSTEMI, the test AUC was 0.94 (95\% CI:0.80-0.90) for the selected model and 0.78 (95\% CI:0.48-0.51) for the GRACE score. For STEMI, the selected XGB model obtained a sensitivity of 0.94 and a specificity of 0.87, while GRACE achieved 0.35 and 0.95, respectively. For NSTEMI, the sensitivity of the selected model was 0.83, and its specificity was 0.87, while the GRACE score was 0.1 and 0.98, respectively. However, GRACE is calculated with only 8 features collected at admission; whereas, the ML-based models use hundreds of features gathered within the first 24 hours of admission. We present the performance of all the trained models in Appendixes C and D.

\subsection{Risk markers in women and men with STEMI and NSTEMI}
\begin{figure*}[!h]
\centering
\subcaptionbox{STEMI}{\includegraphics[scale=0.34]{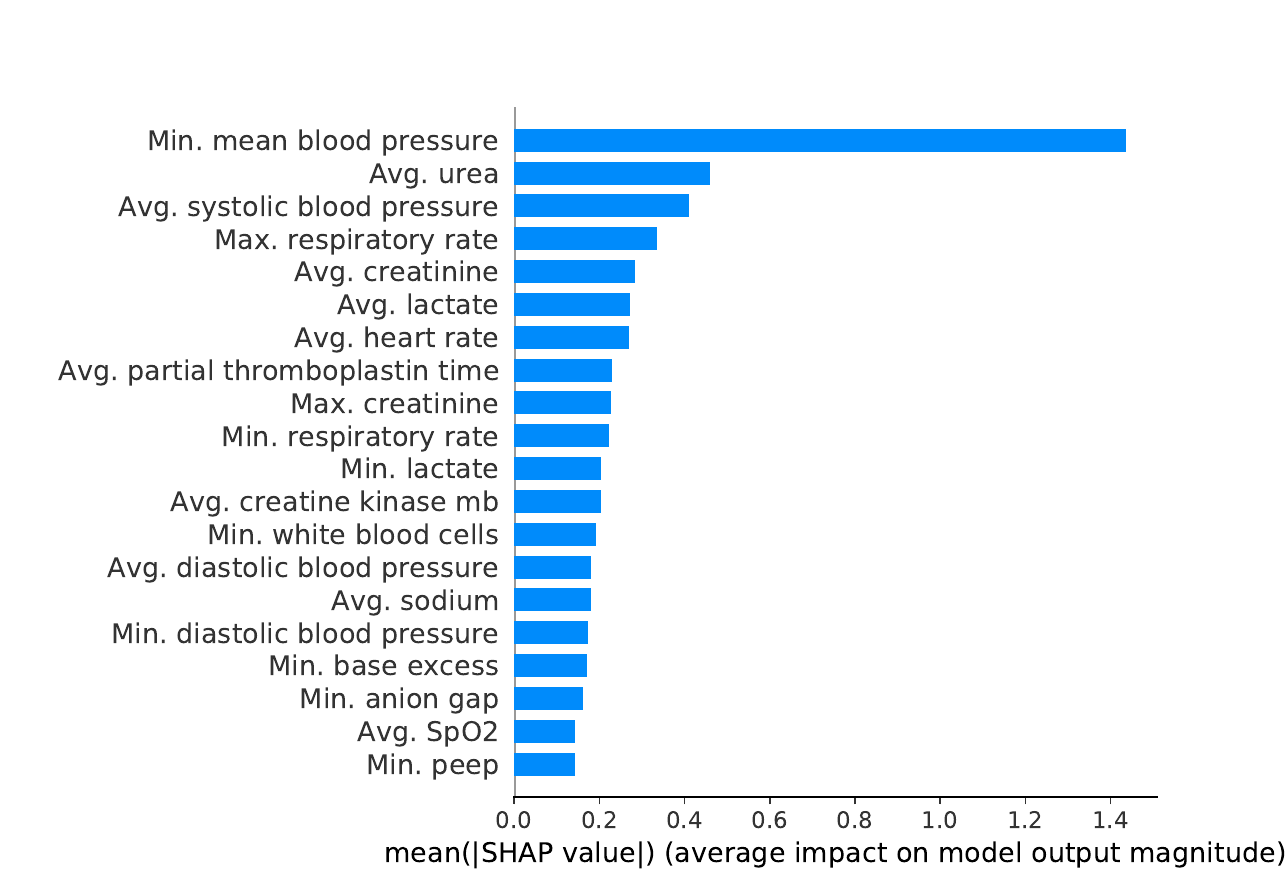}\, {\includegraphics[scale=0.34]{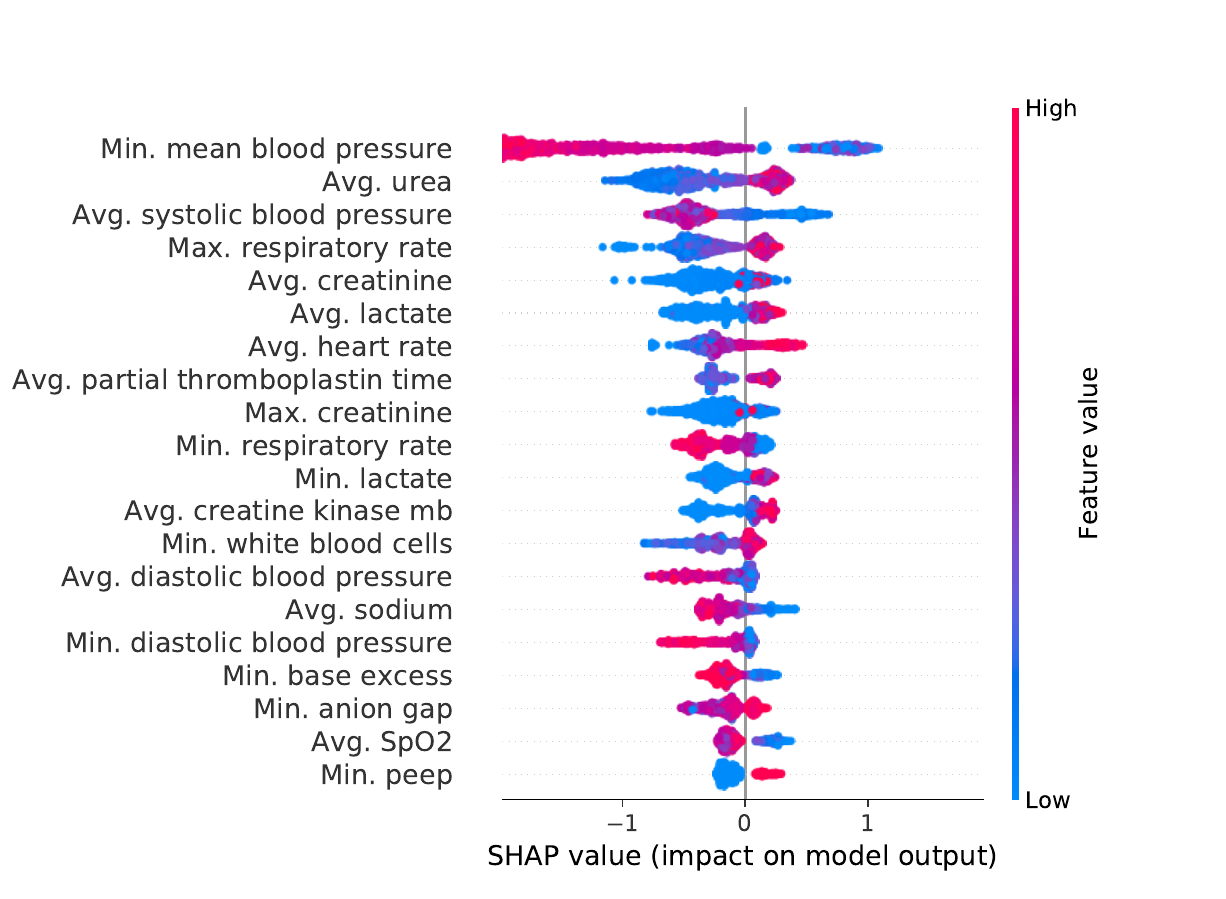}}}
\subcaptionbox{NSTEMI}{\includegraphics[scale=0.34]{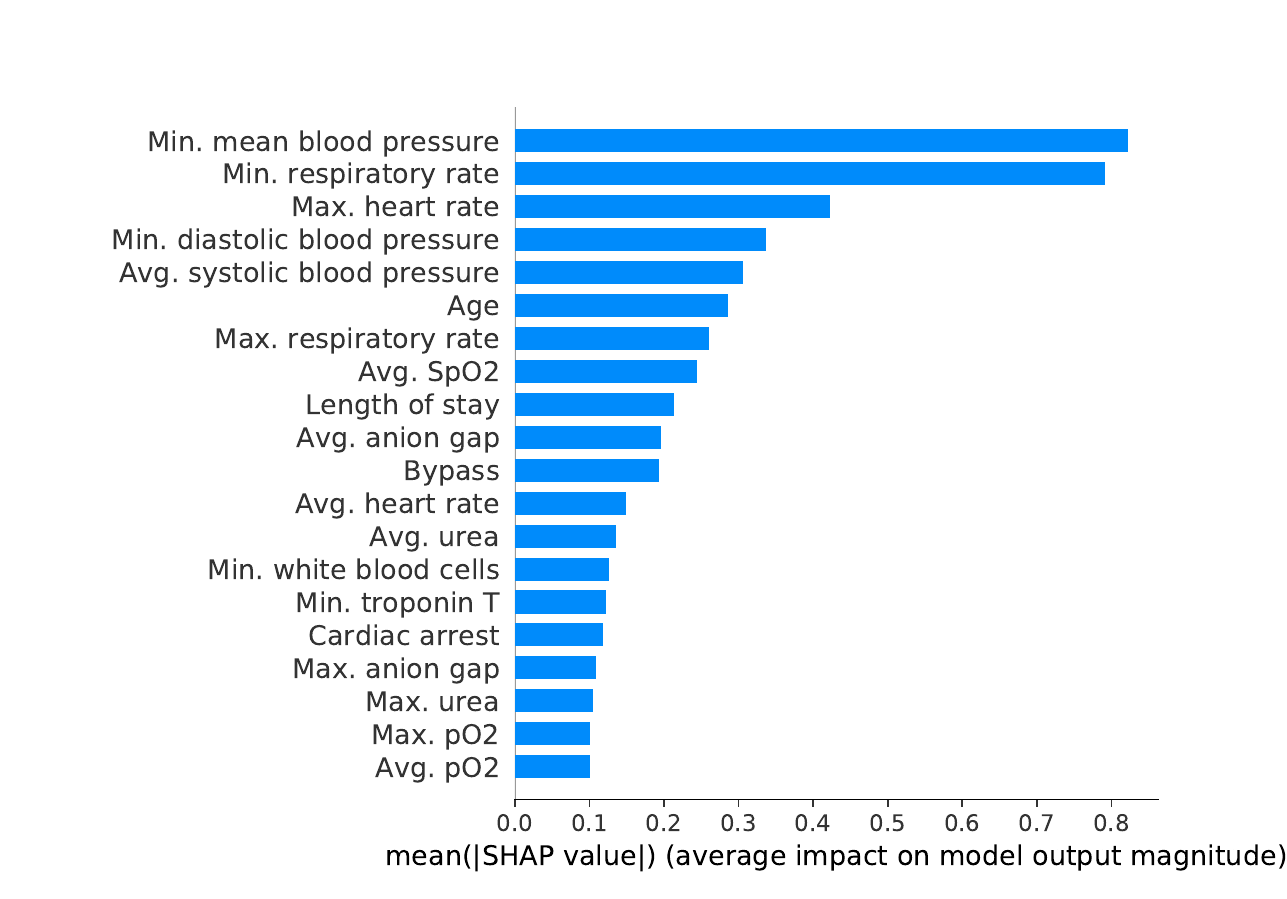}\, {\includegraphics[scale=0.34]{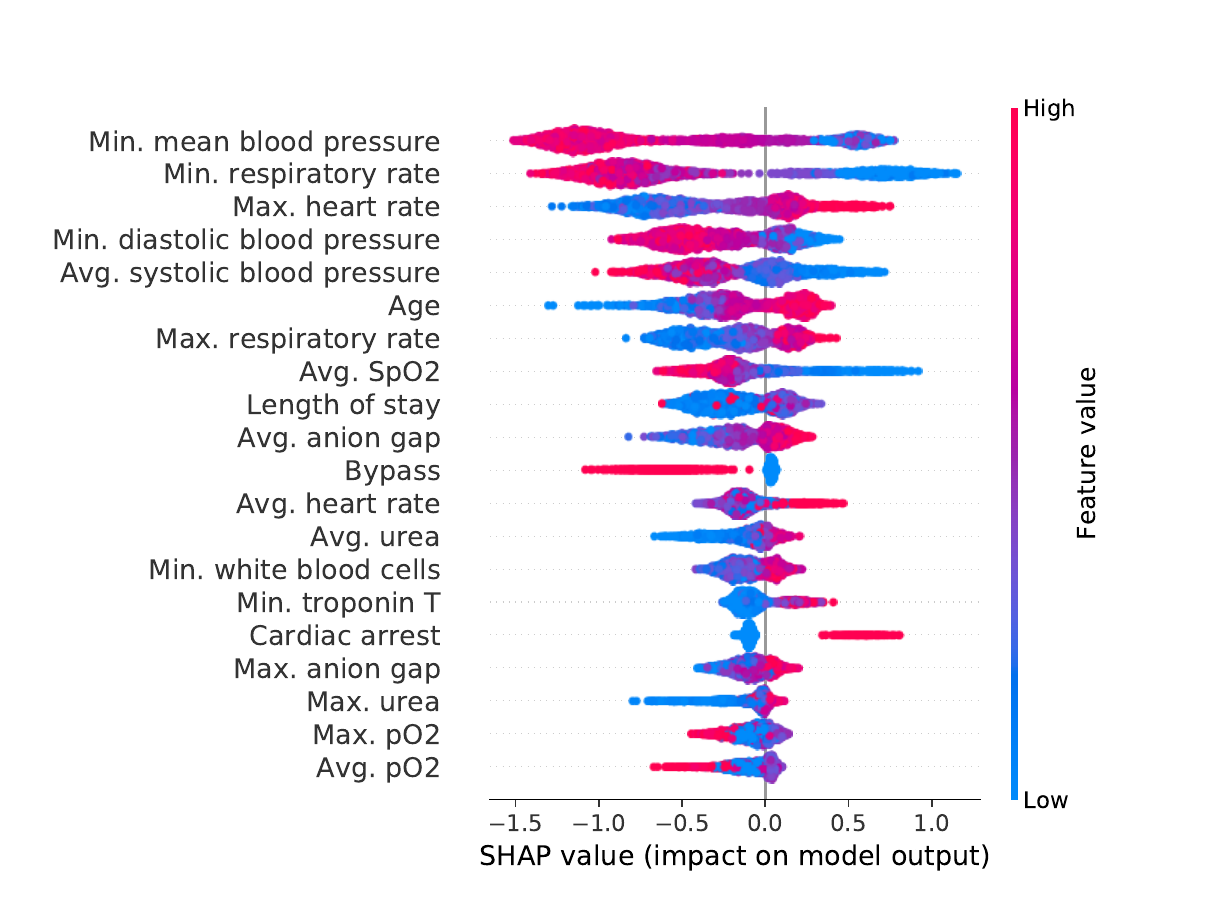}}}
\caption{Top risk markers for STEMI and NSTEMI according to the SHAP approach.}
\label{fig:risk_markers}
\end{figure*}

The XGB models were selected to identify risk markers by computing the SHAP values over the entire dataset of STEMI and NSTEMI. Fig.~\ref{fig:risk_markers} presents the 20 features with the highest SHAP importance (ranked in descending order from top to bottom) for STEMI and NSTEMI, which are considered as the top risk markers. The bar charts on the left of Fig.~\ref{fig:risk_markers} show the SHAP feature importance of these markers. However, the beeswarm plots on the right show the impact on the model’s output for the marker values of individual patients, which are depicted as dots. In the beeswarm plots, larger positive values on the $x$-axis represent a higher mortality risk, whereas, larger negative values refer to a lower risk. Multiple dots with the same $x$-axis form a density. The dot color indicates whether the value of the corresponding feature is high (closer to red) or low (closer to blue).

According to the SHAP approach, common risk markers in both STEMI and NSTEMI (Fig.~\ref{fig:risk_markers}(a) and (Fig.~\ref{fig:risk_markers}(b)) were mean bloop pressure, urea, diastolic blood pressure, systolic blood pressure, respiratory rate, heart rate, and white blood cells. In particular, higher mortality risk is observed with low values for the minimum mean blood pressure and diastolic blood pressure, low values for the average systolic blood pressure, high values for the average urea and heart rate, high values for the minimum white blood cells, and high values for the maximum respiratory rate. 

Specific markers for STEMI were high values of the average creatinine, lactate, partial thromboplastin time, and creatine kinase MB, and low values of the minimum anion gap. In contrast, NSTEMI-specific markers were older with a longer length of stay. NSTEMI-specific markers were also patients who did not undergo bypass surgery, high values of the minimum troponin T, and have a cardiac arrest.

Interestingly, we observed differences between the SHAP feature importances in STEMI and NSTEMI markers. In STEMI, the importance of the minimum mean blood pressure is considerably higher than the rest of the markers. However, in NSTEMI, the minimum mean blood pressure is comparable to the minimum respiratory rate, which is higher than the rest of the markers but not the same as in STEMI. Note that some markers are of the same features. For example, the respiratory rate has markers for the minimum and maximum values in both STEMI and NSTEMI. Creatinine has markers for the average and maximum values in STEMI, and the heart rate has markers for the maximum and average values in NSTEMI. Therefore, it is worth considering the impact of these features on mortality by considering all the associated values.

\begin{figure*}[!h]
\centering
\subcaptionbox{Women with STEMI}{\includegraphics[scale=0.333]{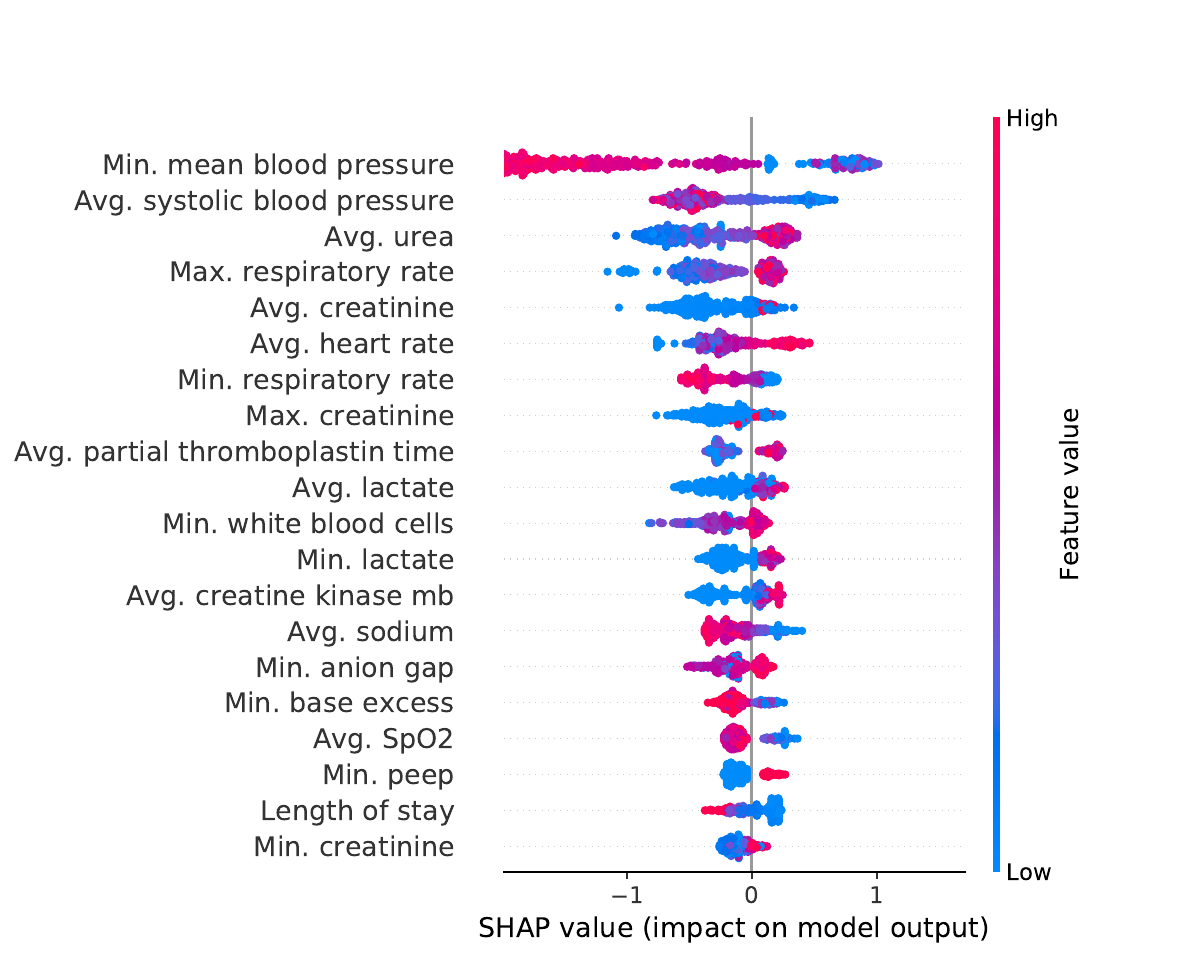}}
\subcaptionbox{Men with STEMI}{\includegraphics[scale=0.333]{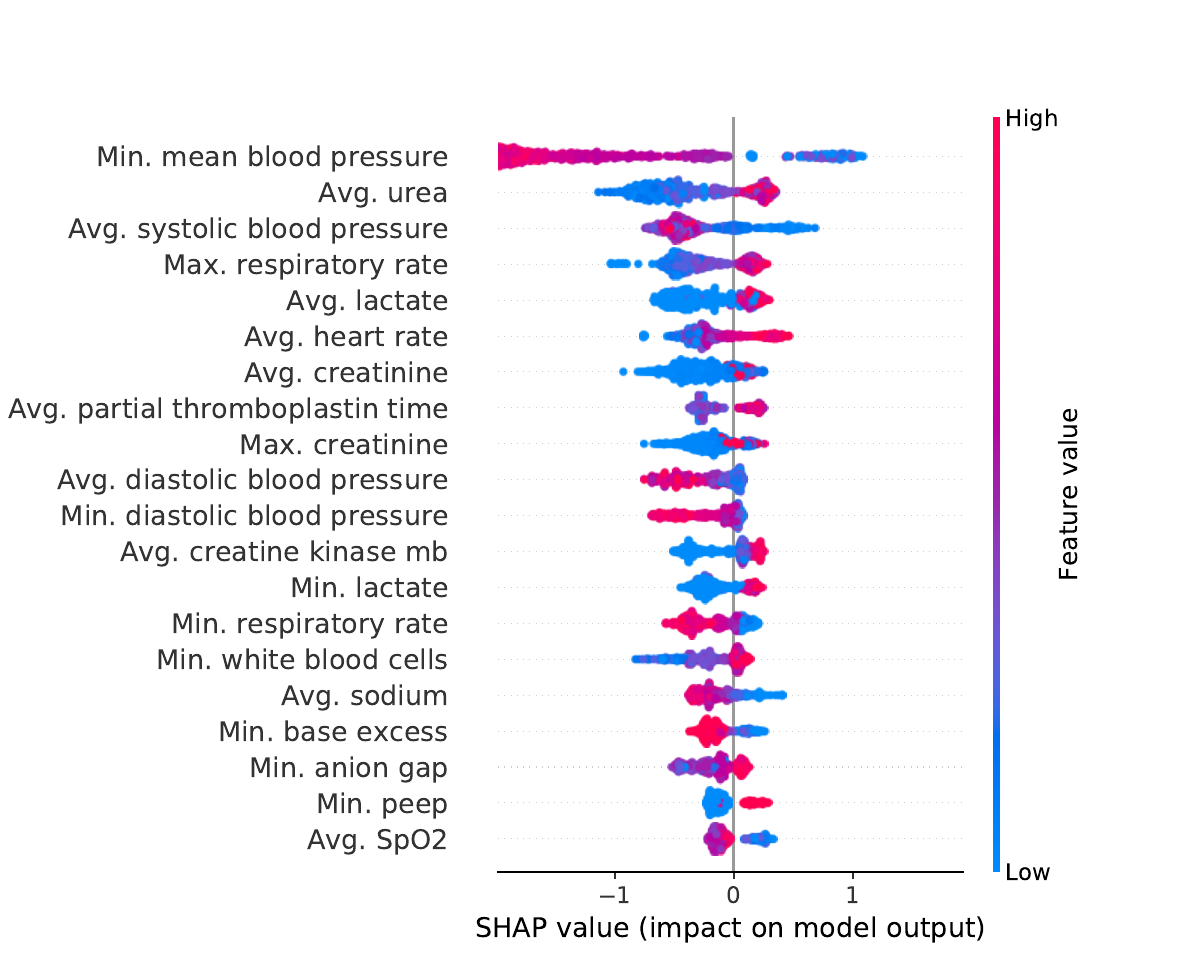}}
\subcaptionbox{Women with NSTEMI}{\includegraphics[scale=0.333]{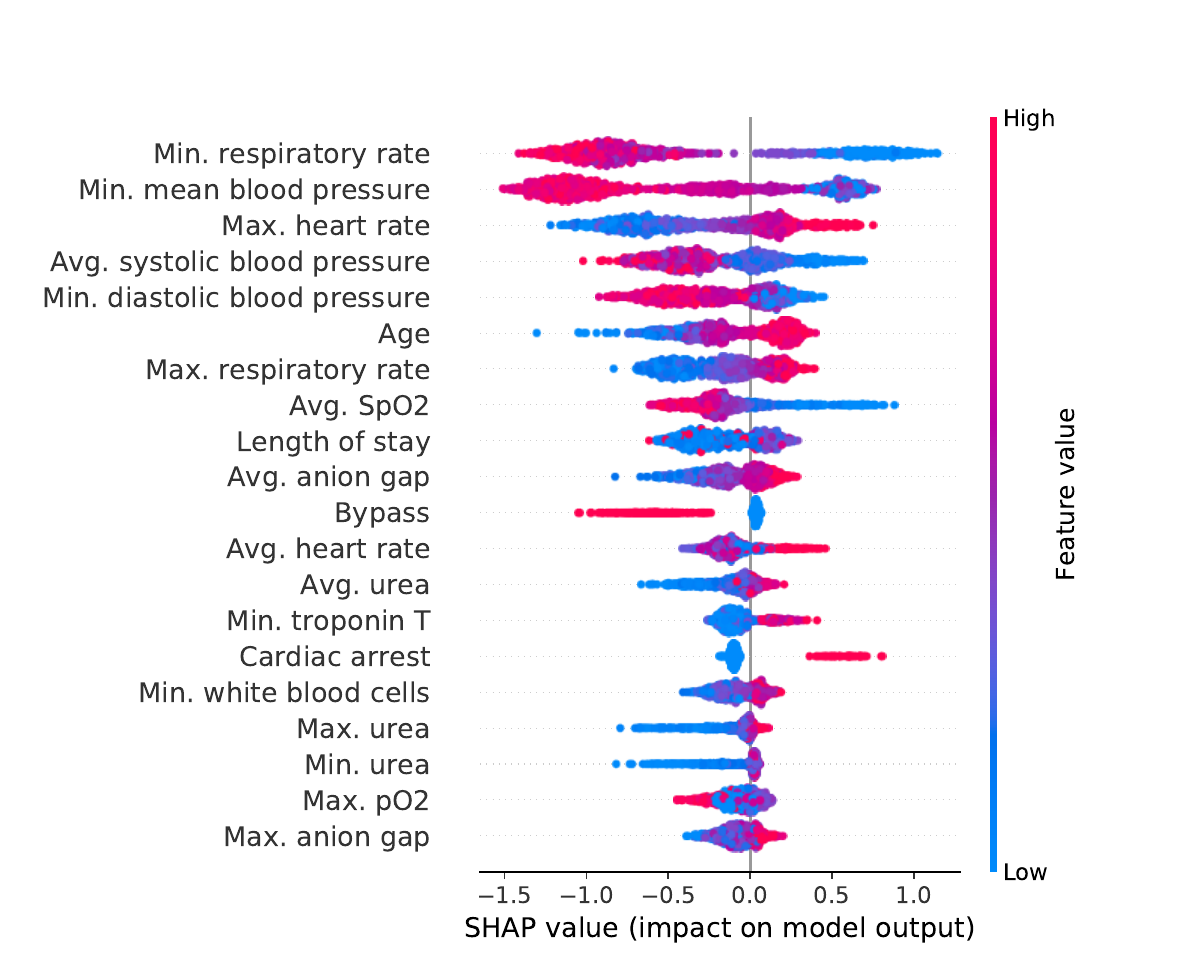}}
\subcaptionbox{Men with NSTEMI}{\includegraphics[scale=0.333]{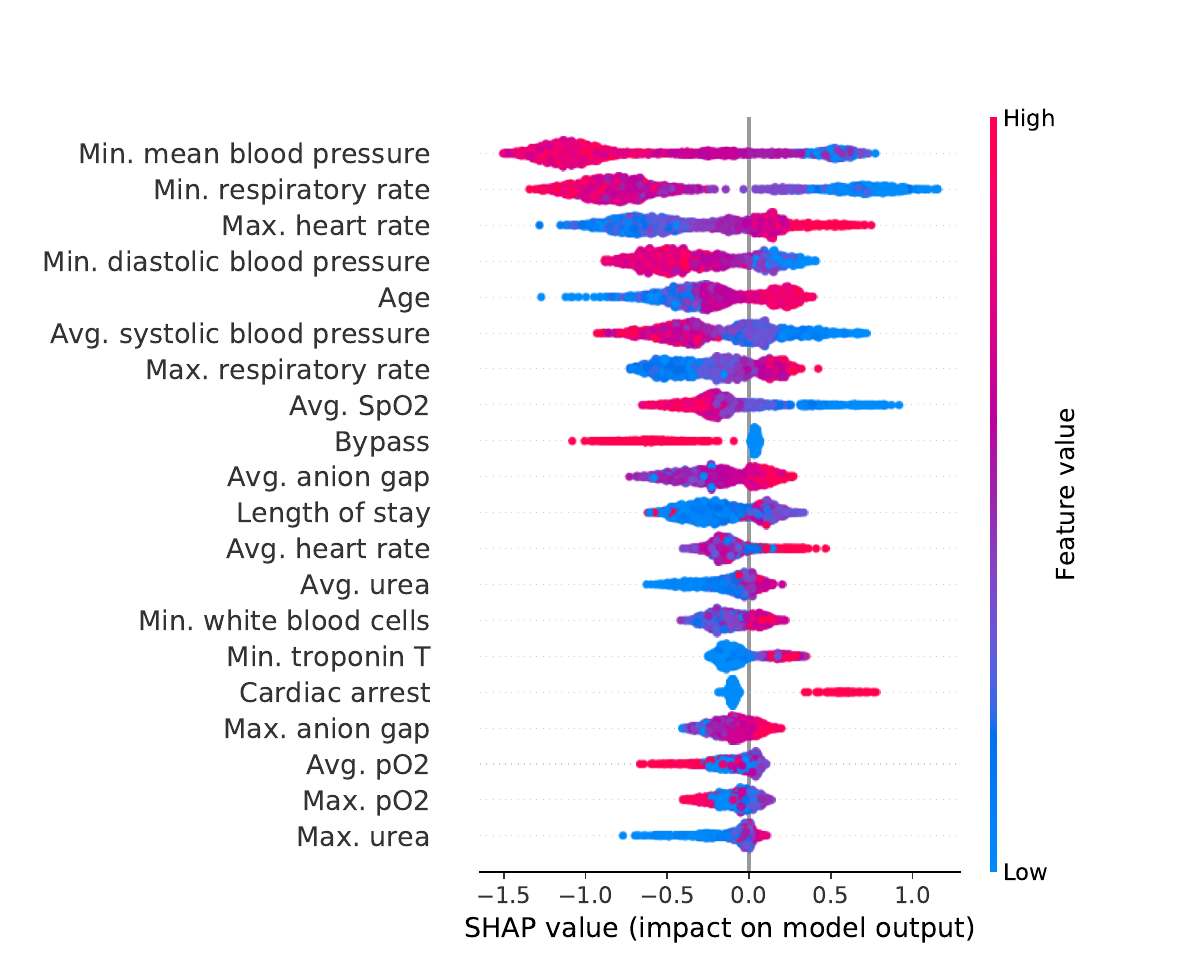}}
\caption{Top sex-specific risk markers for STEMI and NSTEMI.}
\label{fig:specific_sex_markers}
\end{figure*}

To identify sex-specific markers, we generate beeswarm plots with only female patients and only male patients for STEMI and NSTEMI. Fig.~\ref{fig:specific_sex_markers} shows the top sex-specific risk markers for STEMI and NSTEMI. In general, both sub-populations have common markers to the ones identified with all the patients (Fig.~\ref{fig:specific_sex_markers}). The main difference for STEMI is that the average and the minimum diastolic blood pressure are the top risk markers only in men. For NSTEMI, the top markers between women and men are the same.

\begin{figure*}[h!]
\centering
{\includegraphics[scale=0.22]{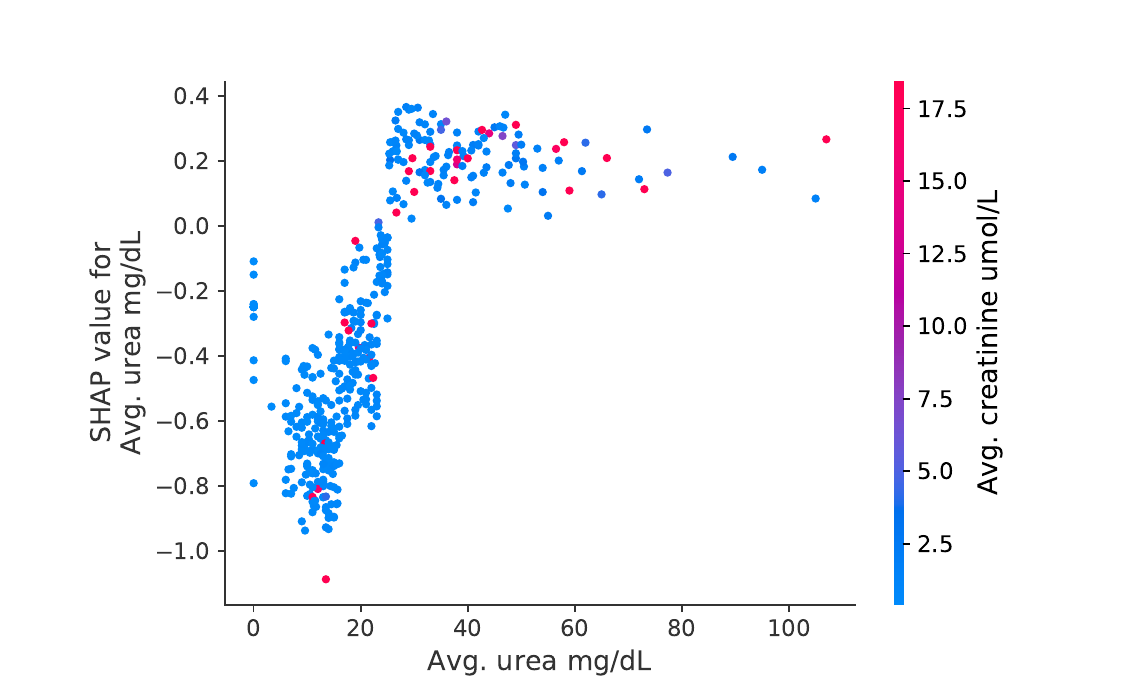}\,
\includegraphics[scale=0.22]{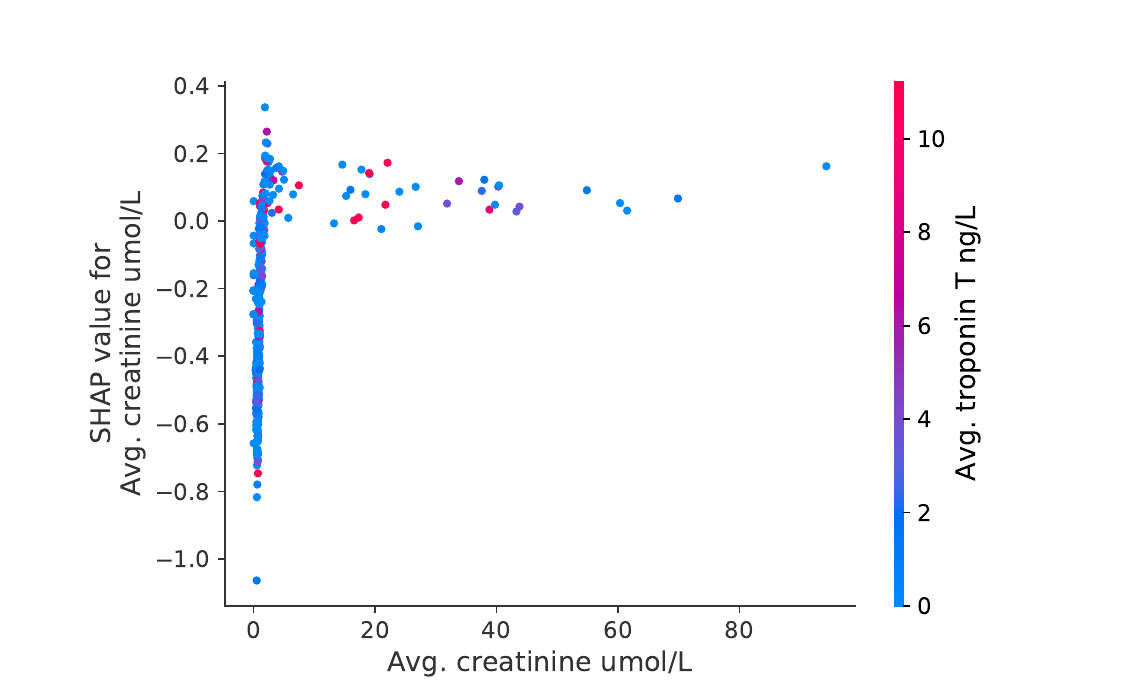}\, \includegraphics[scale=0.22]{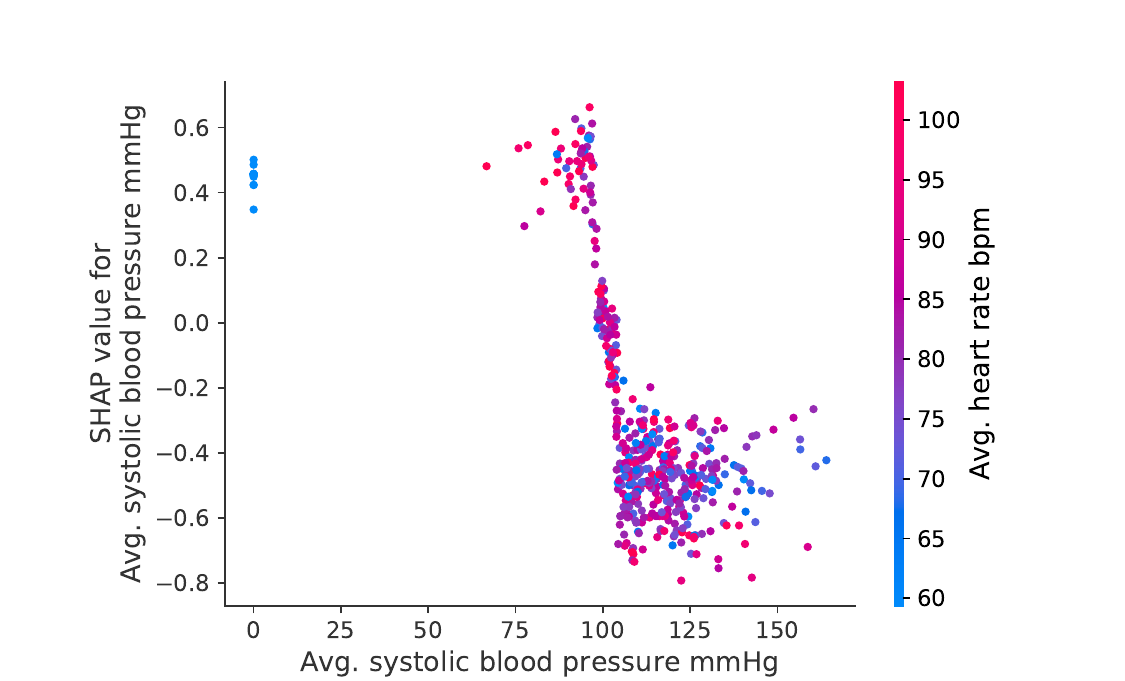}}\\
\subcaptionbox*{Women with STEMI}
{\includegraphics[scale=0.22]{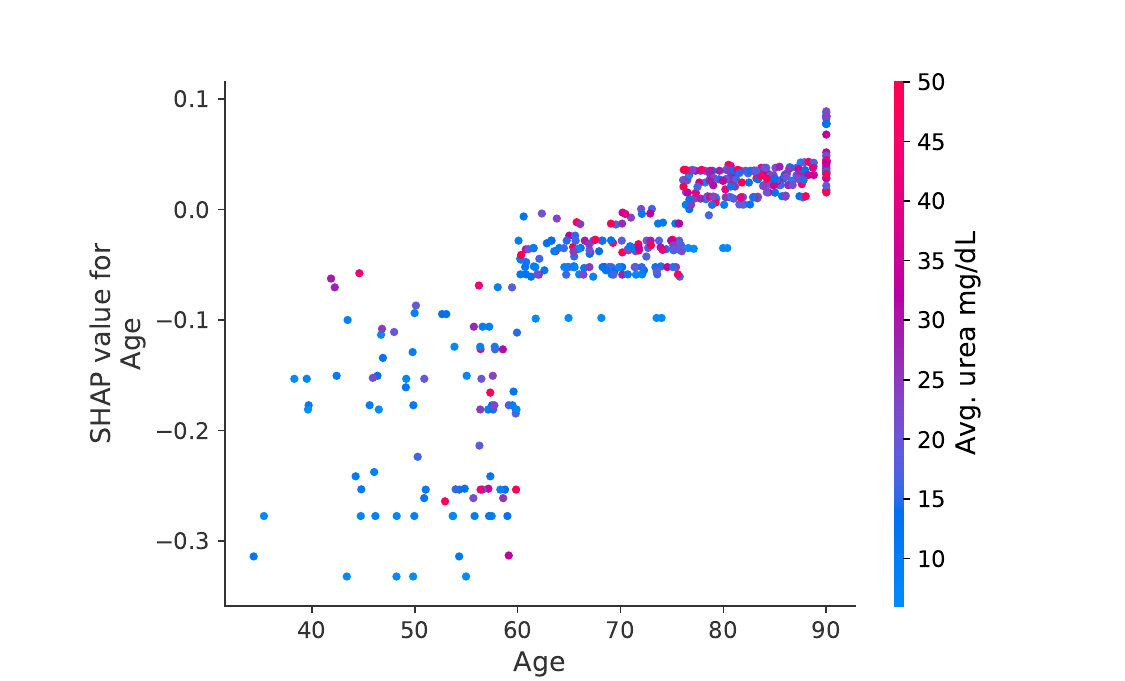}\,
\includegraphics[scale=0.22]{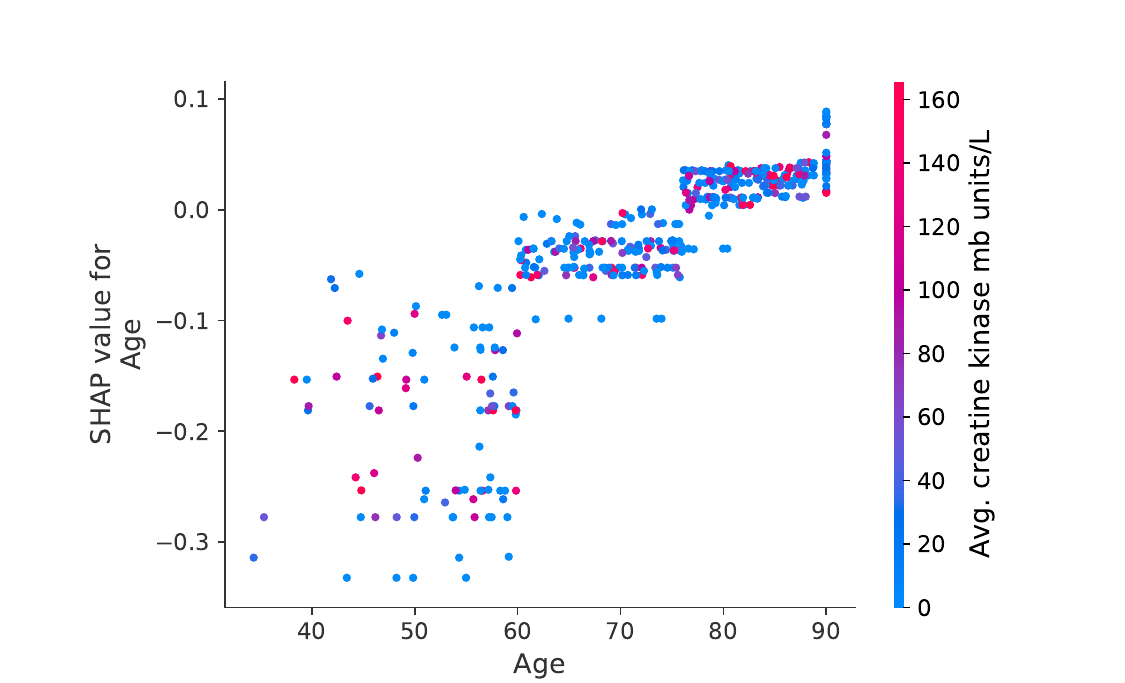}\, \includegraphics[scale=0.22]{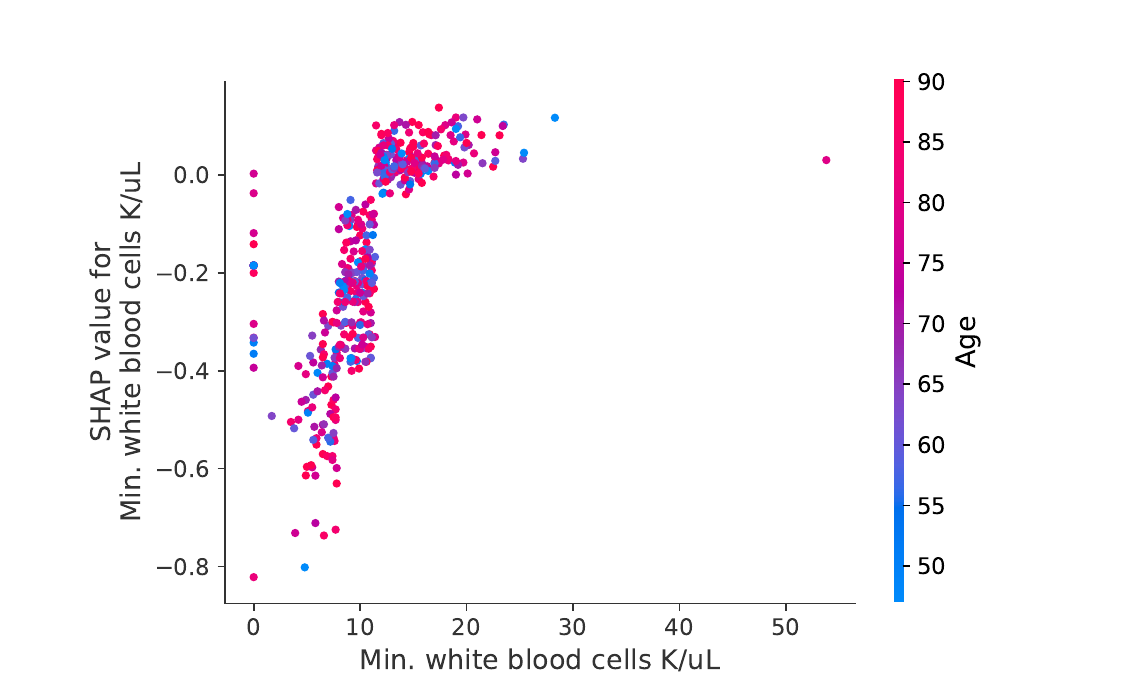}}\\
{\includegraphics[scale=0.22]{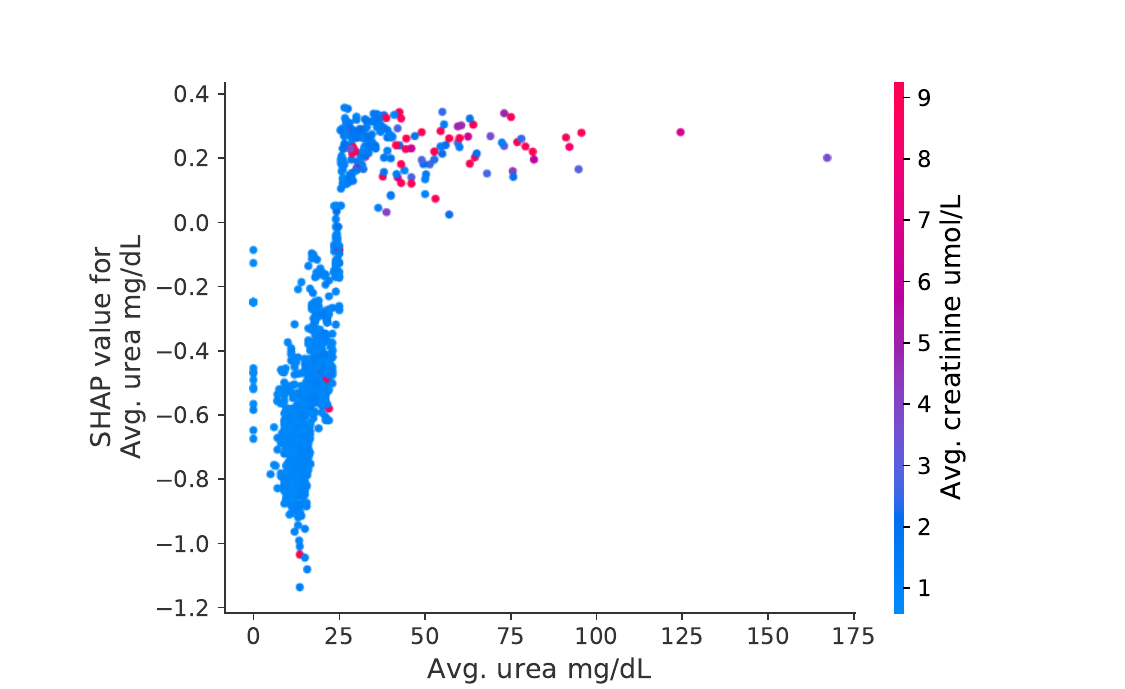}\,
\includegraphics[scale=0.22]{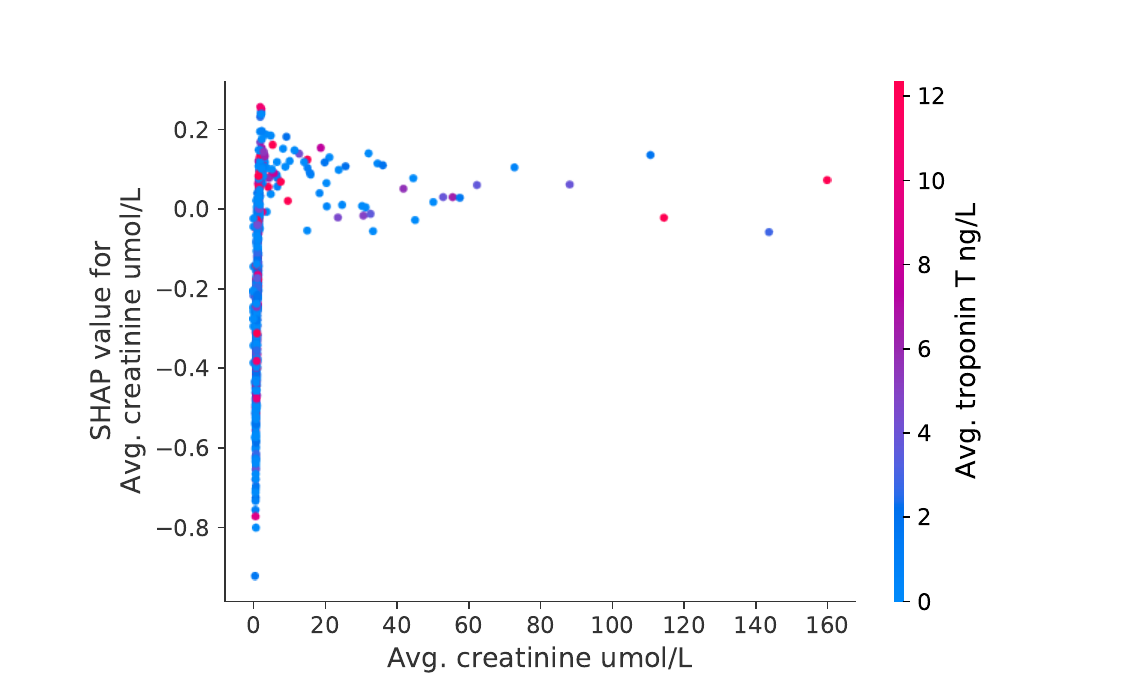}\, \includegraphics[scale=0.22]{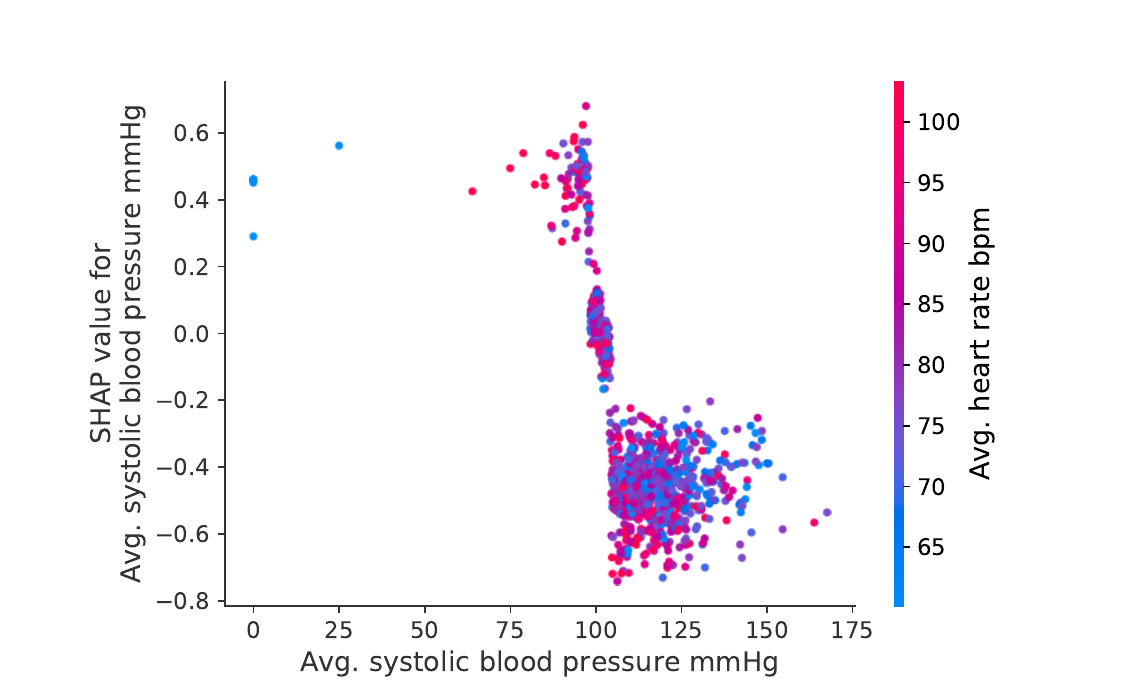}}\\
\subcaptionbox*{Men with STEMI}
{\includegraphics[scale=0.22]{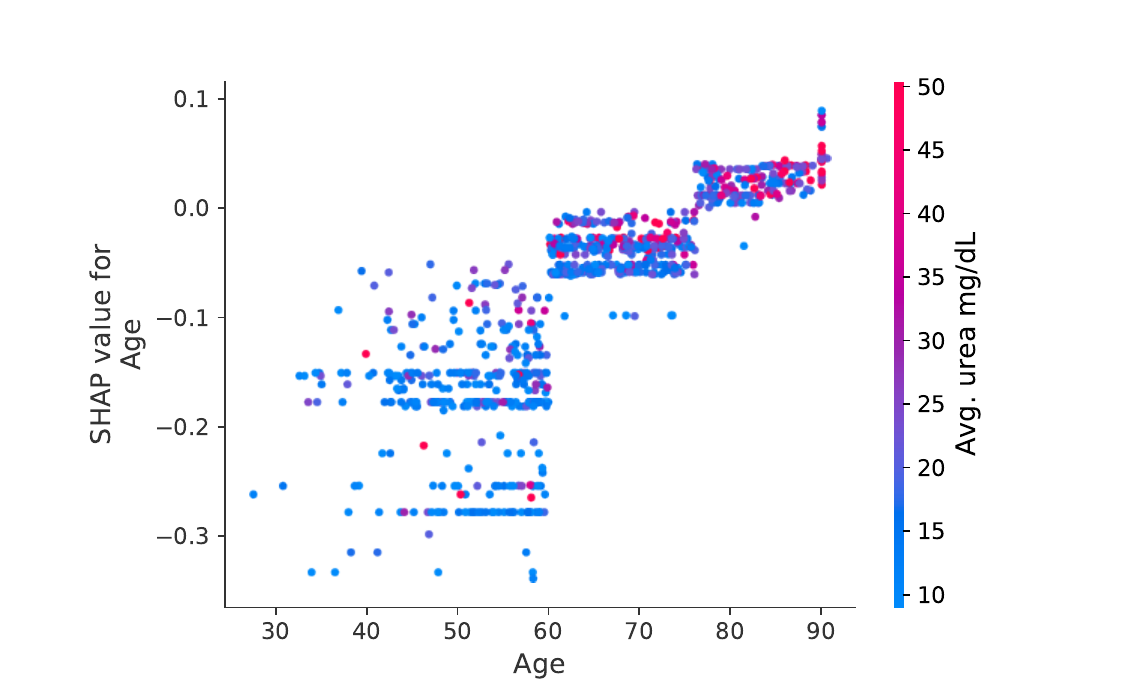}\,
\includegraphics[scale=0.22]{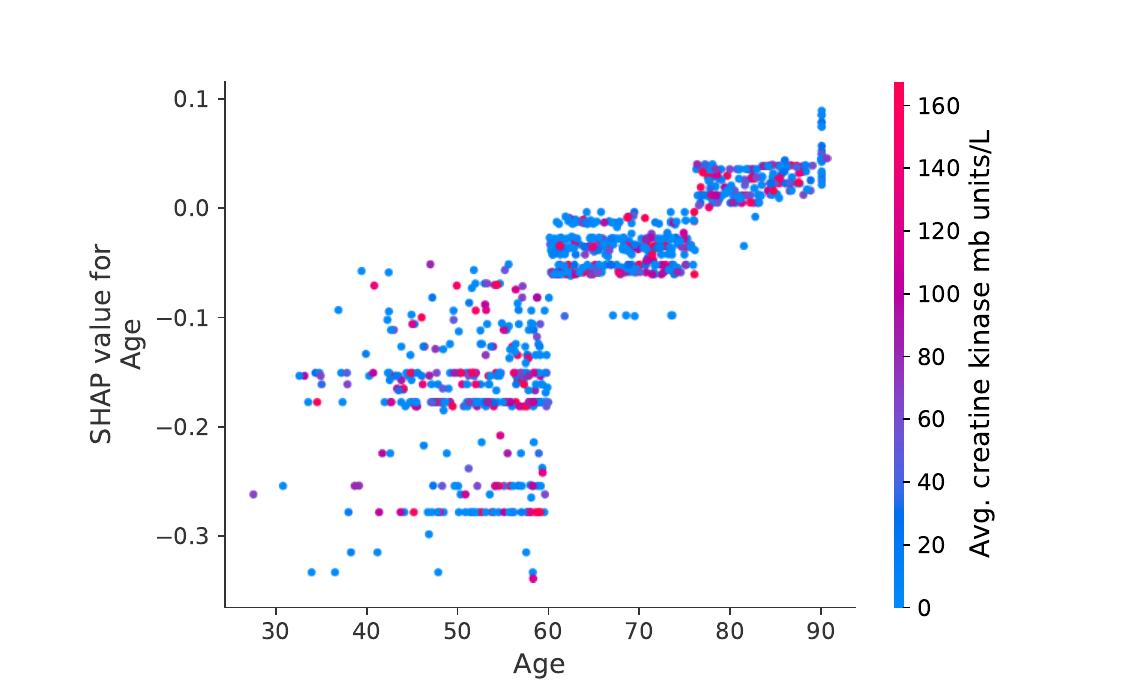}\, \includegraphics[scale=0.22]{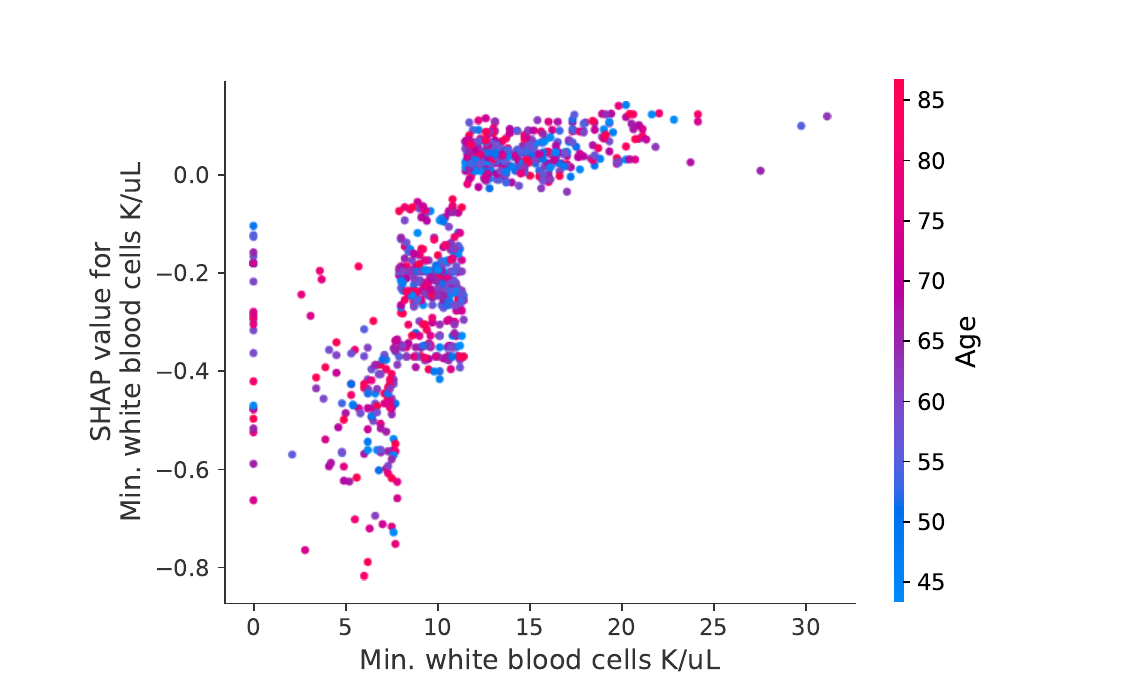}}\\
\caption{The SHAP dependence scatter plots of the identified risk markers in women and men with STEMI.}

\label{fig:markers_stemi_interactions}
\end{figure*}

Moreover, we investigate the sex differences in risk markers by analyzing a set of clinically relevant features selected by expert cardiologists, namely age, urea, creatinine, troponin T, creatine kinase MB, heart rate, white blood cells, and mean and systolic blood pressure. We generate the SHAP dependence scatter plots to analyze the impact of a selected feature on the model’s output and its relation with other relevant features. The $x$-axis in a SHAP dependence scatter plot represents the range of values for the selected feature, and the $y$-axis represents the range of SHAP values for the same feature. Larger positive SHAP values indicate a higher risk, and larger negative SHAP values indicate lower risk. As revealed in beeswarm plots, each dot is an individual patient, and its color denotes the value of another relevant feature, with lower values closer to blue and higher values closer to red.

Fig.~\ref{fig:markers_stemi_interactions} presents the SHAP scatter plots for women and men with STEMI. From the plots, high values of the average urea increase the risk in patients with high values of the average creatinine. Note that the values of the average urea are higher in men than women. However, the values of the average creatinine are higher in women than men. Similarly, high values of the average creatinine increase the risk in the patient suffering from high values of the average troponin T. In this case, the creatinine levels and troponin T are higher in men than women. We also found that the average urea and creatine kinase MB, as well as the minimum white blood cells, have a higher impact in elder patients. Increased risk in women and men with STEMI and NSTEMI are low values of the average systolic blood pressure and high values of the average heart rate.

\begin{figure}[h!]
\centering
{\includegraphics[scale=0.22]{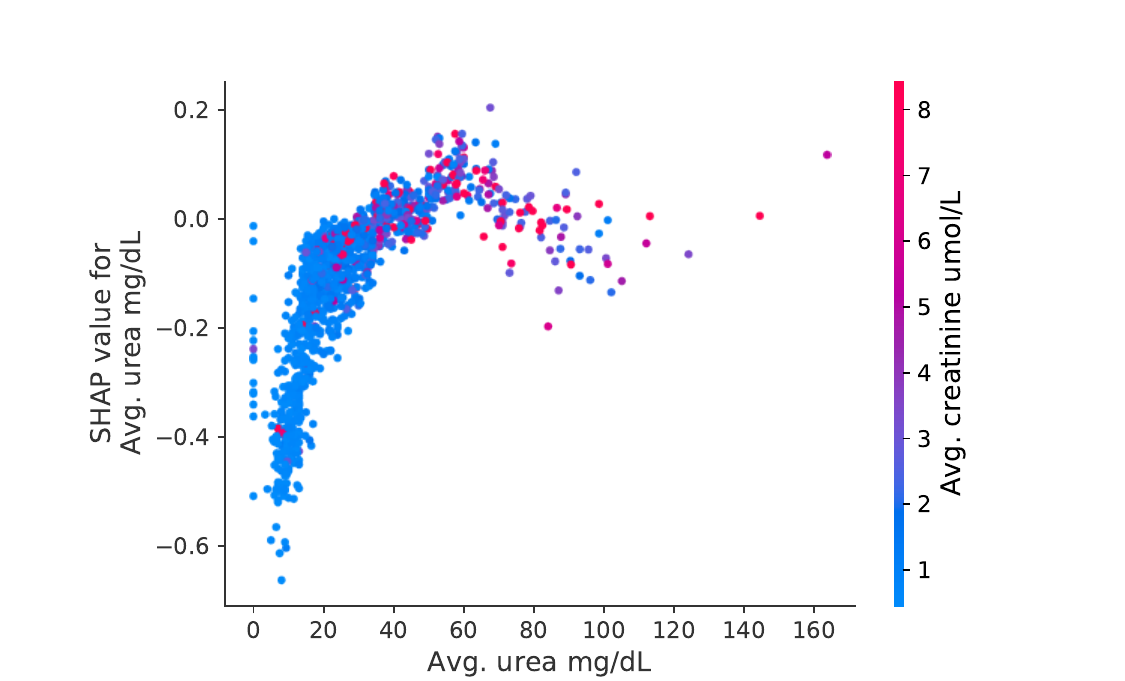}\,
\includegraphics[scale=0.22]{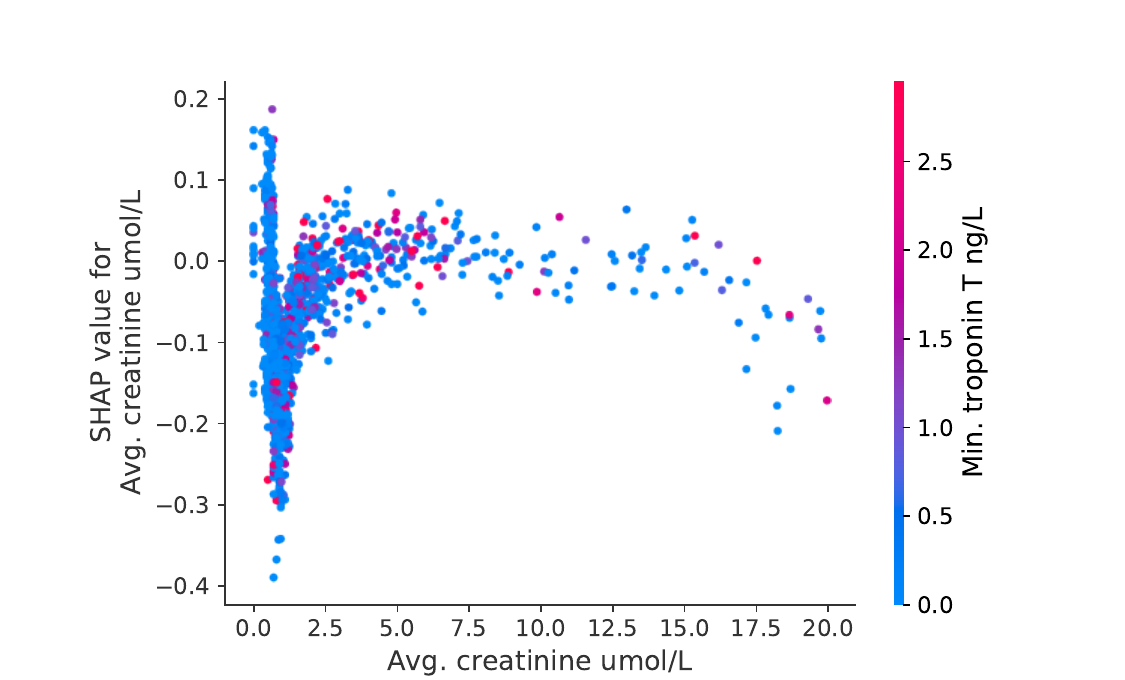}\, \includegraphics[scale=0.22]{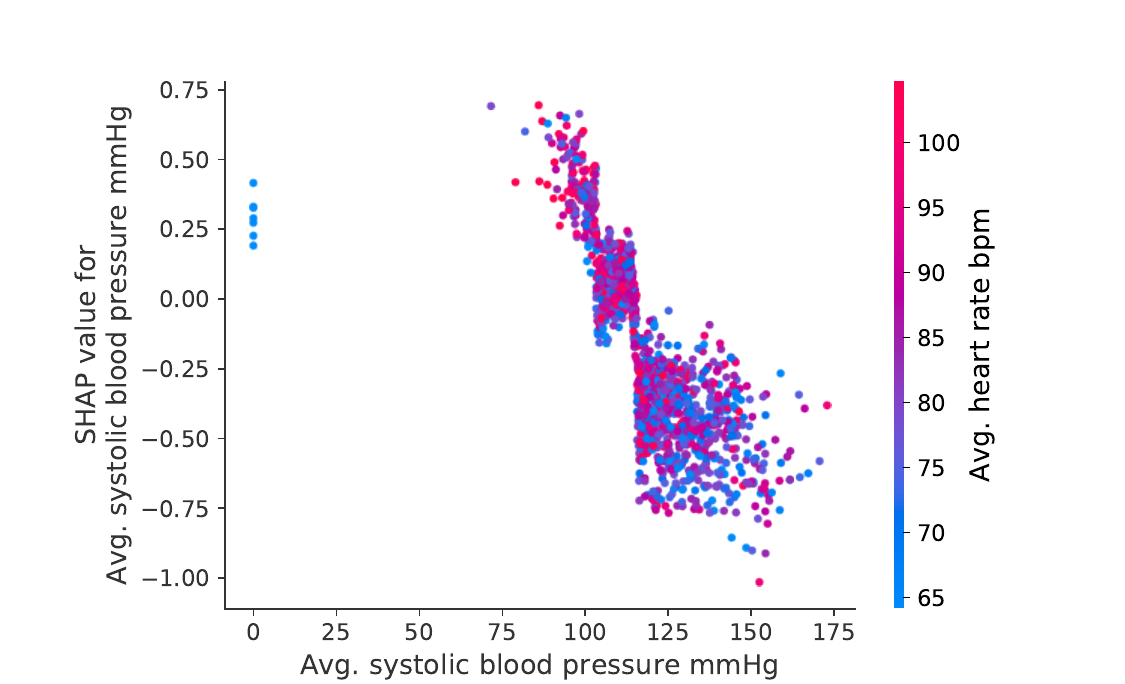}}\\
{\includegraphics[scale=0.22]{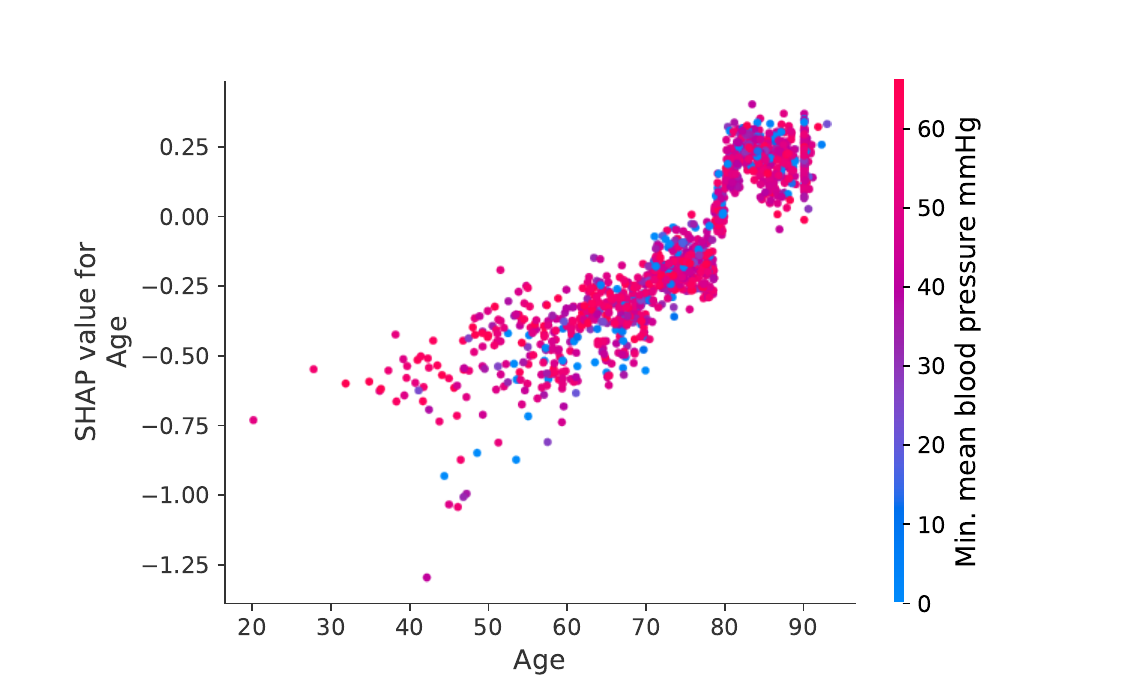}\,
\includegraphics[scale=0.22]{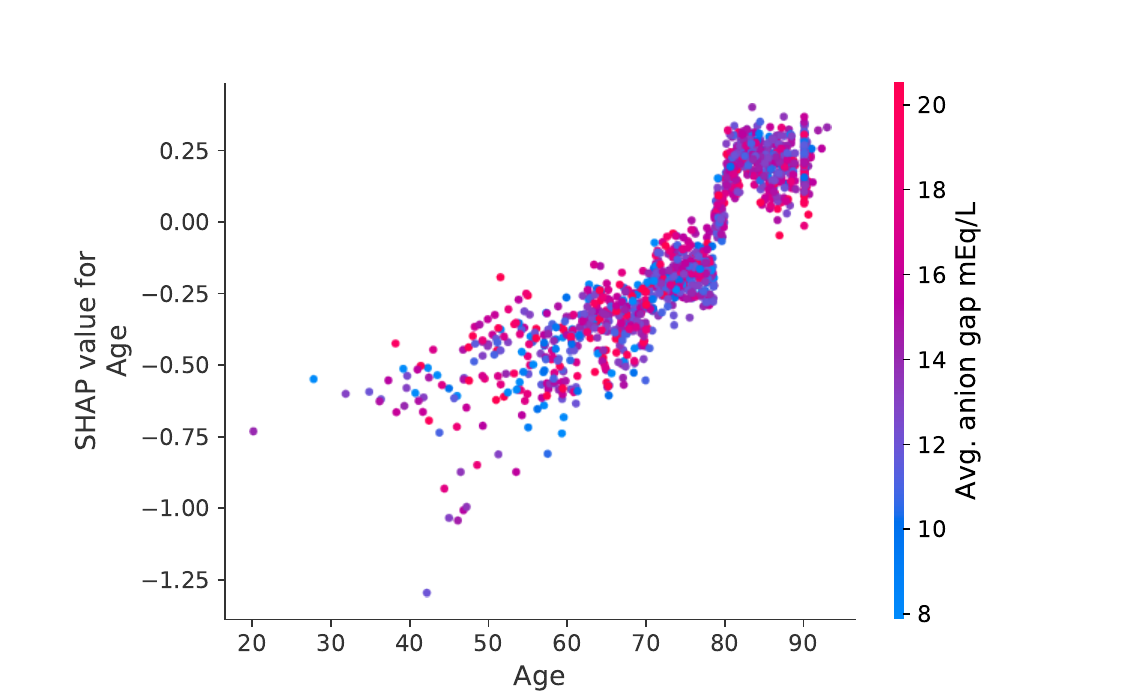}\, \includegraphics[scale=0.22]{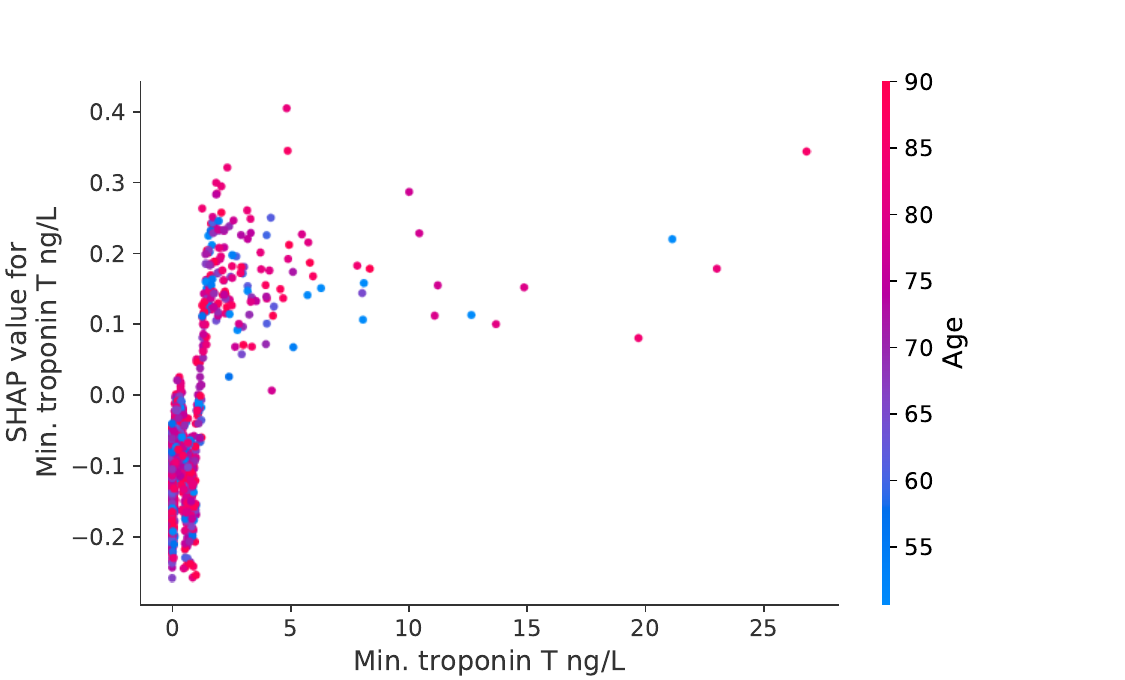}}\\
\caption*{Women with NSTEMI}
{\includegraphics[scale=0.22]{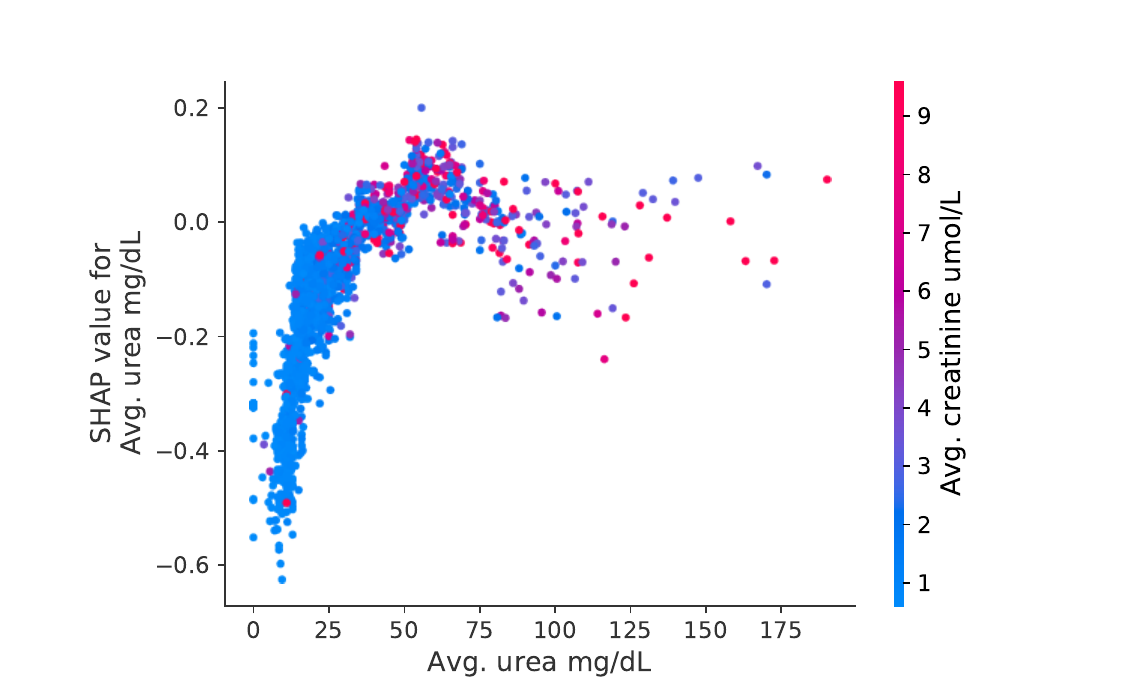}\,
\includegraphics[scale=0.22]{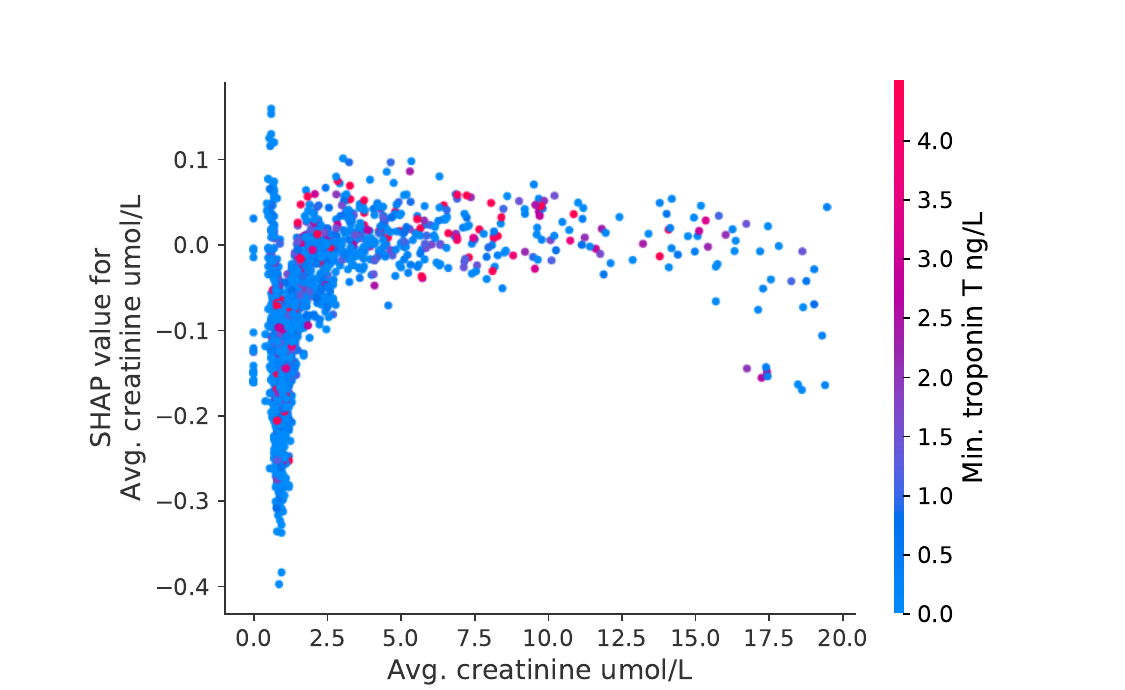}\, \includegraphics[scale=0.22]{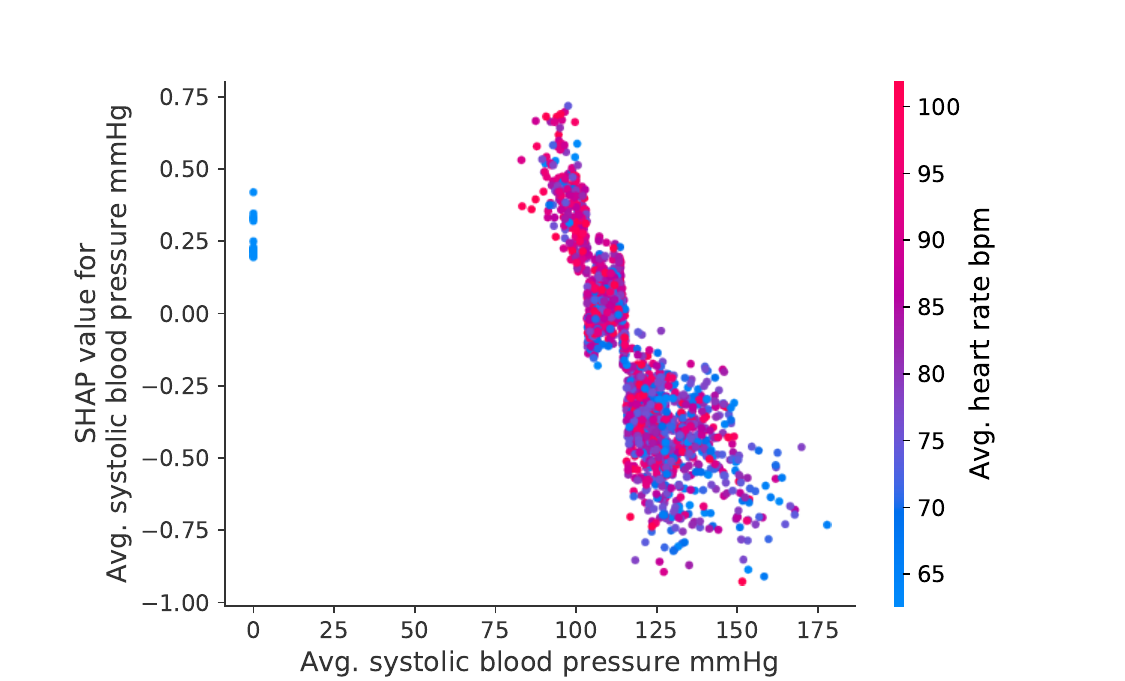}}\\
{\includegraphics[scale=0.22]{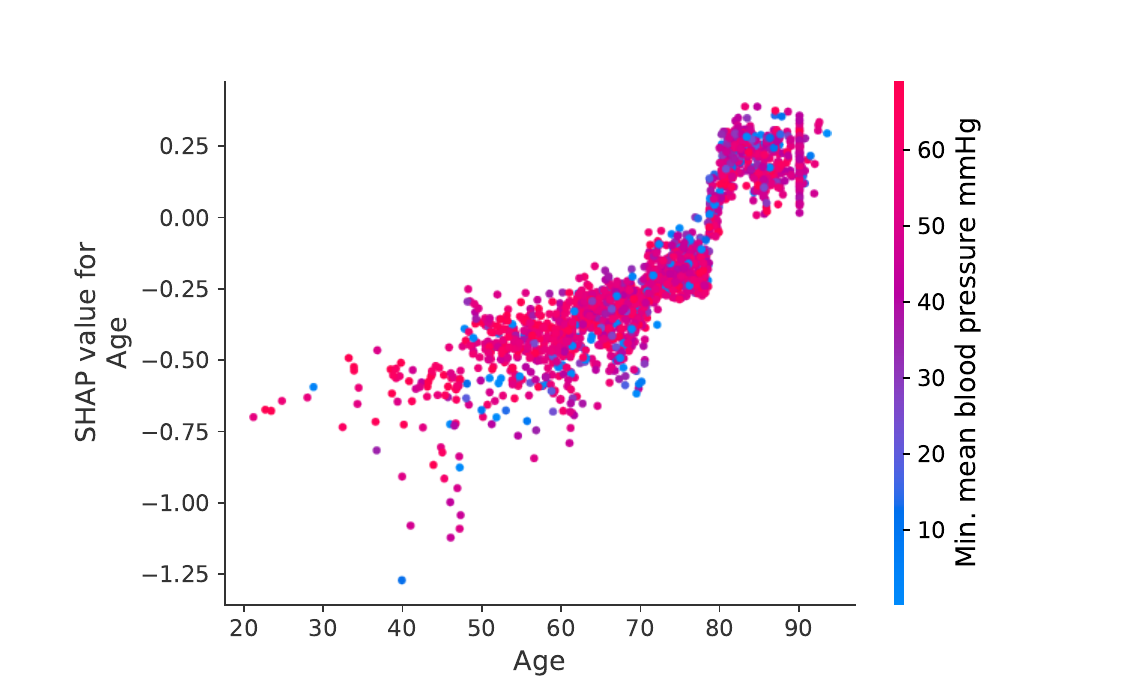}\,
\includegraphics[scale=0.22]{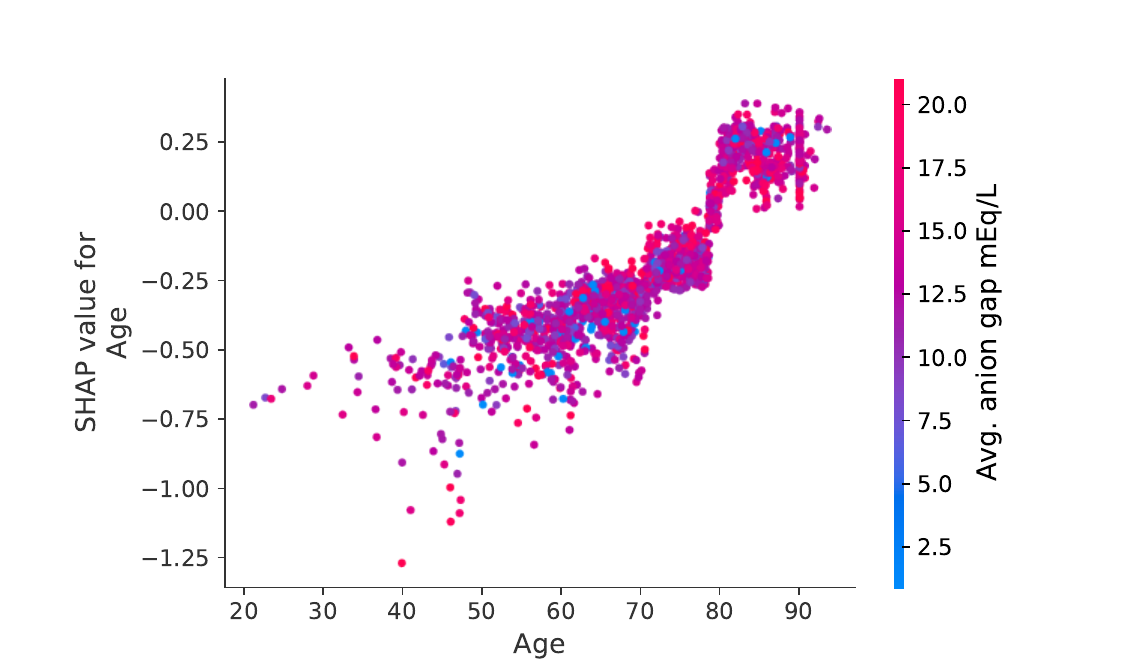}\, \includegraphics[scale=0.22]{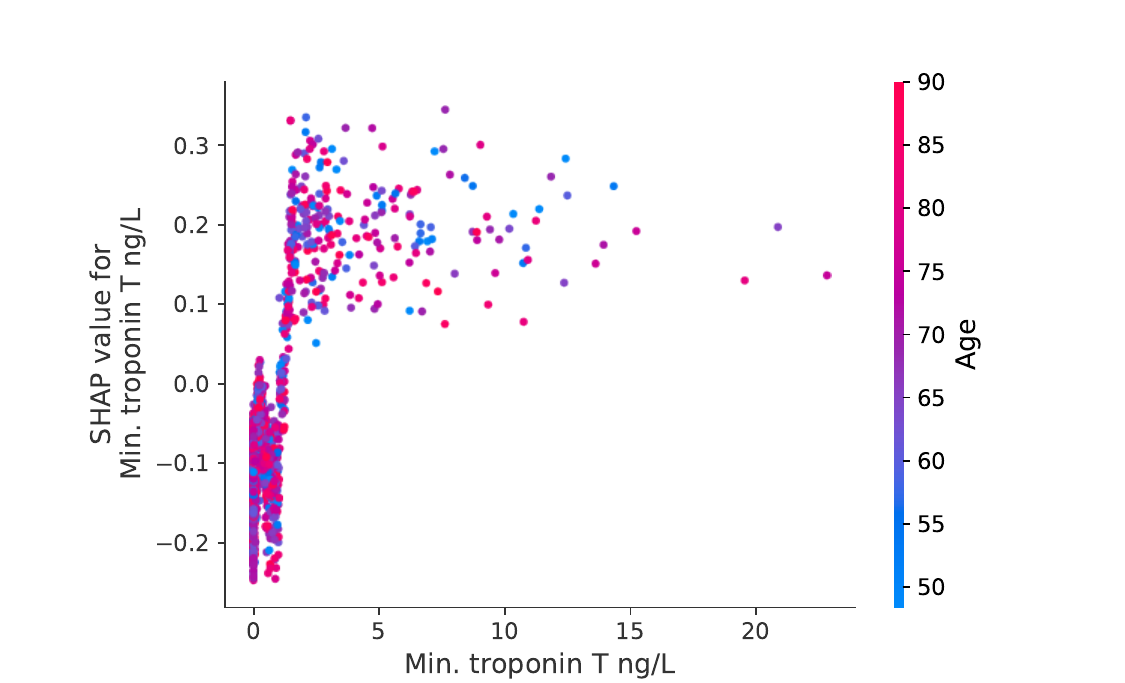}}\\
\caption*{Men with NSTEMI}
\caption{The SHAP dependence scatter plots of the identified risk markers in women and men with NSTEMI.}
\label{fig:markers_nstemi_interactions}
\end{figure}

Fig.~\ref{fig:markers_nstemi_interactions} displays the SHAP dependence scatter plots for women and men with NSTEMI. Here, patients with high values of the average urea face a higher risk when they have high values of the average creatinine. For NSTEMI as opposed to STEMI, the values of the average urea are higher in women than in men. However, high values of the average creatinine show a higher impact on mortality rate when there are high values of the minimum troponin T. Note that men have higher values of the minimum troponin levels than women. Finally, we found that high levels of the minimum blood pressure, the average anion gap, and minimum troponin T increase the risk of older patients.

As revealed in Fig.~\ref{fig:markers_stemi_interactions} and Fig.~\ref{fig:markers_nstemi_interactions}, we identified a set of critical levels for the selected features using SHAP values above zero as a threshold, which were summarized in Table~\ref{tab:interaction_markers}. Here, an increase in the risk is the STEMI average urea levels above 20 mg/dL in women and 25 mg/dL in men along with the average creatinine levels under 17.5 umol/L in women and 9 umol/L in men. However, markers that increase the risk in women are the NSTEMI with higher average urea levels than 30 mg/dL with lower average creatinine levels than 8 umol/L. On the other hand, markers that augment the risk in men are higher average urea levels than 25 mg/dL and lower average creatinine levels than 9 umol/L. Generally, women over 70 years have a higher risk in STEMI than men over 75 years. Women over 80 years have a higher risk of NSTEMI than men over 70 years. Although these critical levels provide concrete information about sex differences in risk markers, extensive research is still required. 

\begin{table}[!ht]
\caption{\label{tab:interaction_markers} Summary of the sex differences in the identified risk markers for STEMI and NSTEMI.}
\begin{center}
\scalebox{0.47}{
\begin{tabular}{lll|lll}
\hline
\multicolumn{3}{c|}{\textbf{STEMI}}
& \multicolumn{3}{c}{\textbf{NSTEMI}}
\\ \hline
\multicolumn{1}{c}{All}&
\multicolumn{1}{c}{Women}&
\multicolumn{1}{c|}{Men}&
\multicolumn{1}{c}{All} &
\multicolumn{1}{c}{Women} &
\multicolumn{1}{c}{Men}\\ \hline

\begin{tabular}[c]{@{}l@{}}
Prolonged \\ thromboplastin times \end{tabular}
&
\begin{tabular}[c]{@{}l@{}}
Urea \textgreater{} 20 mg/dL and \\creatinine \textless{} 17.5 umol/L\end{tabular}
&
\begin{tabular}[c]{@{}l@{}}
Urea \textgreater{} 25 mg/dL and \\ creatinine \textless{} 9 umol/L\end{tabular}
&
\begin{tabular}[c]{@{}l@{}}
Low values of sysbp\\and diasbp\end{tabular}
&
\begin{tabular}[c]{@{}l@{}}
Urea \textgreater{} 30 mg/dL and \\ creatinine \textless{} 8 umol/L\end{tabular}
&
\begin{tabular}[c]{@{}l@{}}
Urea \textgreater{} 25 mg/dL and\\ creatinine \textless{} 9 umol/L\end{tabular}

\\ \hline
\begin{tabular}[c]{@{}l@{}}
Low values of\\ sodium\end{tabular}
&
\begin{tabular}[c]{@{}l@{}}
Creatinine \textgreater{} 4 umol/L and \\ troponin-T \textless{} 10 ng/L\end{tabular}
&
\begin{tabular}[c]{@{}l@{}}
Creatinine \textgreater{} 5 umol/L and\\ troponin-T \textless{} 12 ng/L\end{tabular}
&
\begin{tabular}[c]{@{}l@{}}
Long of stays\end{tabular}
&
\begin{tabular}[c]{@{}l@{}}
Creatinine \textgreater{} 1.5 umol/L and\\troponin-T \textless{} 2.5 ng/L\end{tabular}
&
\begin{tabular}[c]{@{}l@{}}
Creatinine \textgreater{} 1.5 umol/L and\\ troponin \textless{} 4 ng/L\end{tabular}

\\ \hline
\begin{tabular}[c]{@{}l@{}}
High values of\\ PEEP \end{tabular}
&
\begin{tabular}[c]{@{}l@{}}
Sysbp \textless{} 110 mmHg \\ and heart rate \textgreater{} 80\end{tabular}
&
\begin{tabular}[c]{@{}l@{}}
Systolic bp \textless{} 110 mmHg \\ and heart rate \textgreater{} 70\end{tabular}
&
\begin{tabular}[c]{@{}l@{}}
Cardiac arrest\end{tabular}
&
\begin{tabular}[c]{@{}l@{}}
Systolic bp \textless{} 120 mmHg \\ and heart rate \textgreater{} 75\end{tabular}
&
\begin{tabular}[c]{@{}l@{}}
Systolic bp \textless{} 120 mmHg \\ and heart rate \textgreater{} 85\end{tabular}
\\ \hline
\begin{tabular}[c]{@{}l@{}}
Low values of\\ base excess\end{tabular}
&
Age \textgreater{} 70 years
&
Age \textgreater{} 75 years
&
\begin{tabular}[c]{@{}l@{}}
High values of\\heart rate\end{tabular}
&
Age \textgreater{} 80 years
&
Age \textgreater{} 70 years
\\ \hline
\end{tabular}
}
\end{center}
\tiny `All' refers to both women and men; diasbp: diastolic blood pressure; PEEP: Positive End-Expiratory Pressure; sysbp: systolic blood pressure.
\end{table}

\subsection{Individual predictions for patients with STEMI and NSTEMI}

Using SHAP Waterfall plots, we also examined how individual feature values contribute to the model’s output. These plots show the output value $f(x)$ for a given instance in the $x$-axis, along with the average output value of all the patients $E[f(X)]$ as reference. The rows in the $y$-axis are the most important features (ranked in descending order from top to bottom). Their corresponding SHAP values are depicted as red or blue arrows of different lengths, which push the output to the left or right of $E[f(X)]$ over the $x$-axis and increase or decrease the model’s output value. Red arrows push the output towards a higher value (higher risk). However, blue arrows push the output towards a lower value (lower risk). Here, the direction and length of the arrows represent the direction and the magnitude of the contribution of the corresponding feature value to the output. In addition, the features with the smallest contributions are combined in the bottom row of the plot.

\begin{figure*}[h!]
\centering
\subcaptionbox{Woman with a high risk}
{\includegraphics[scale=0.24]{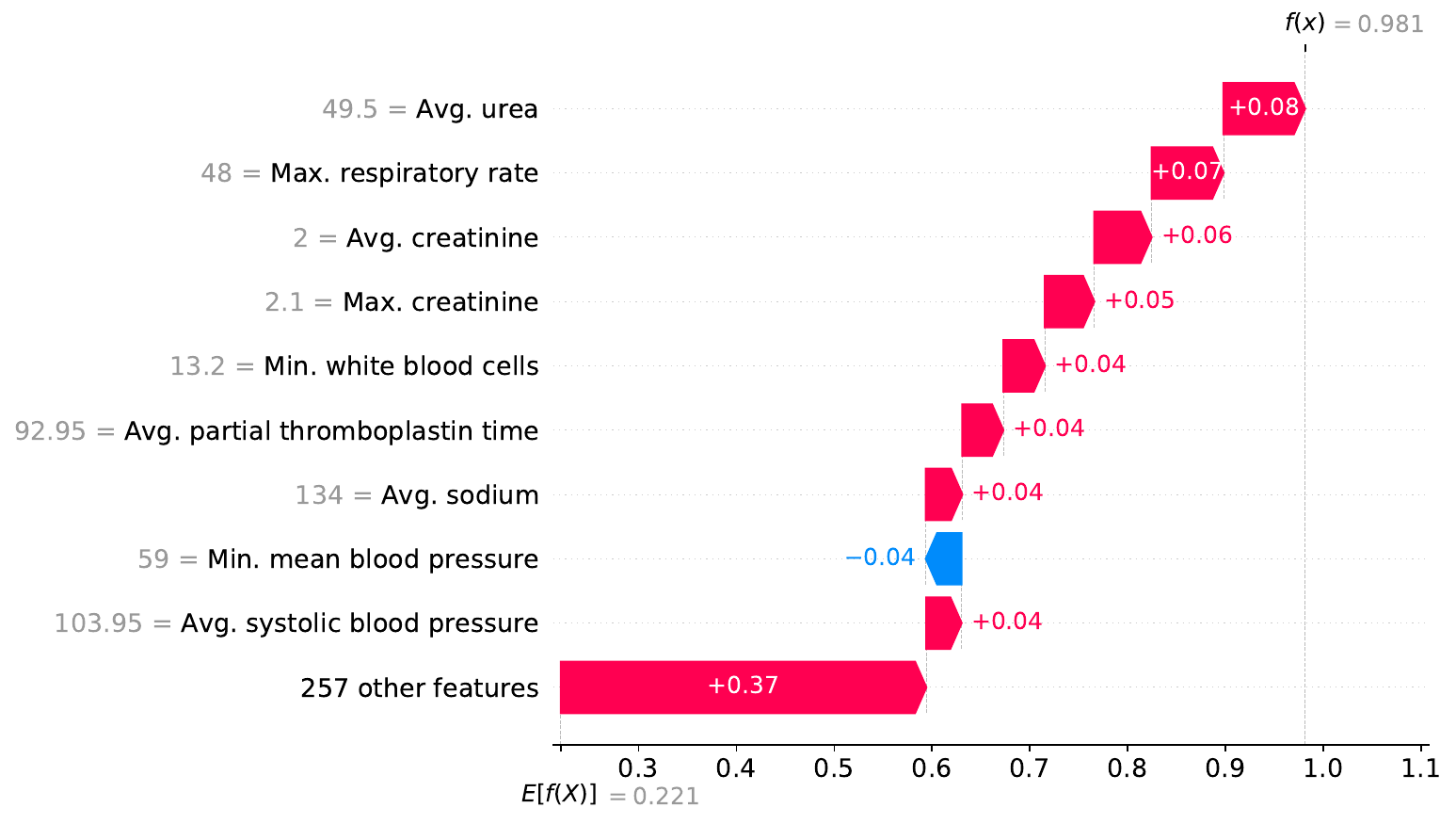}} 
\subcaptionbox{Woman with a low risk}
{\includegraphics[scale=0.24]{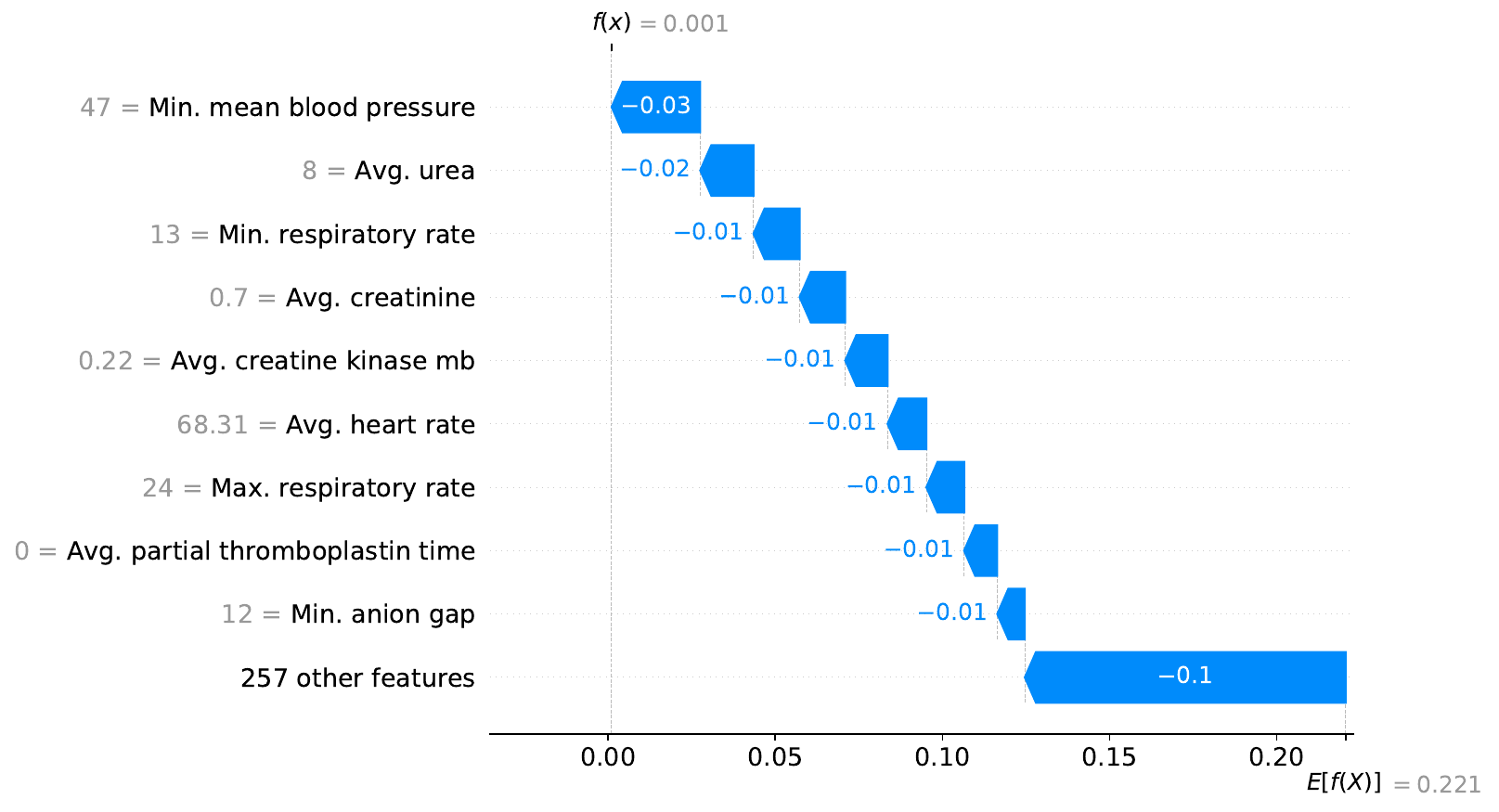}}
\subcaptionbox{Man with a high risk}{\includegraphics[scale=0.24]{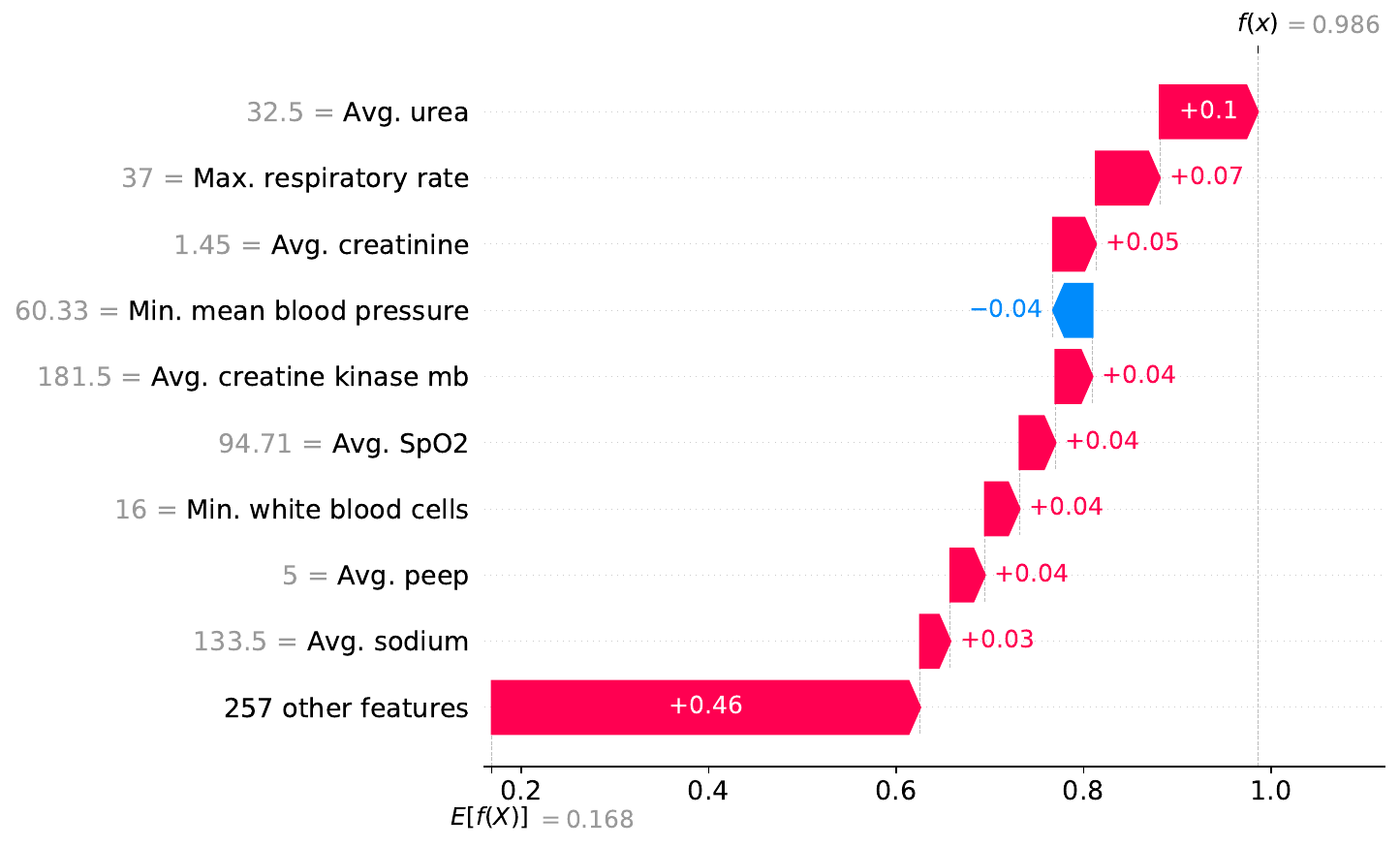}}
\subcaptionbox{Man with a low risk}{\includegraphics[scale=0.24]{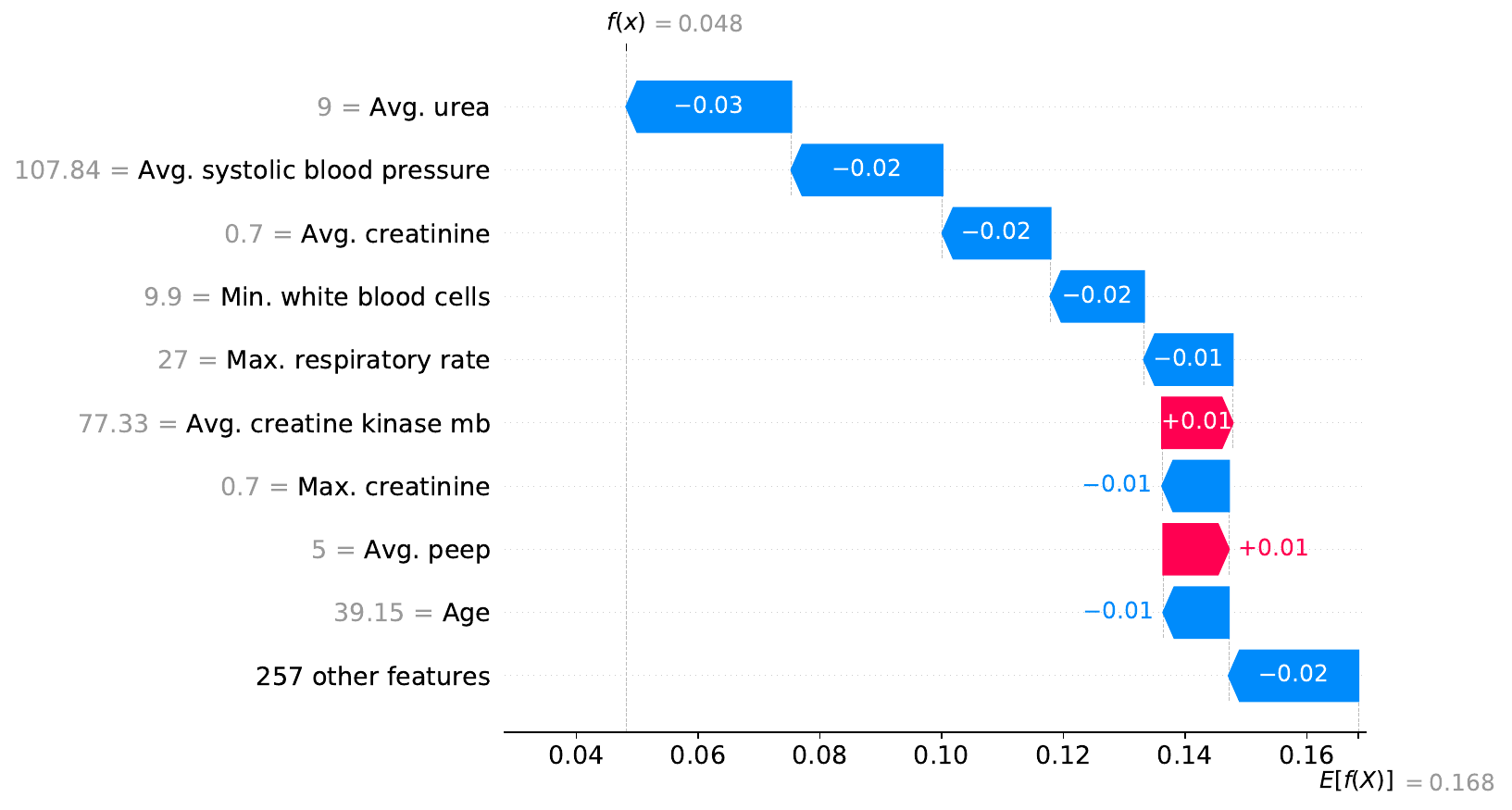}}
\caption{Explanations of individual predictions for patients with STEMI.}
\label{fig:ind_pred_stemi}
\end{figure*}

Fig.~\ref{fig:ind_pred_stemi} presents the waterfall plots for the predictions of four sample patients with STEMI. Fig.~\ref{fig:ind_pred_stemi}(a) depicted the explanation of a woman who died. In this case, the model’s output was a high mortality risk ($f(x) = 0.981$). The feature values that pushed the prediction more strongly towards a higher risk for this patient were high levels of the average urea (49.5 mg/dL), high levels of the maximum respiratory rate (48 pbm), and high values of the average and maximum creatinine levels (2 umol/L and 2.1 umol/L, respectively). Conversely, Fig.~\ref{fig:ind_pred_stemi}(b) illustrates the prediction of a woman who survives. For this patient, the model’s output value was a low probability of mortality ($f(x)=0.001$). The feature values that reduced the probability of mortality were values of the minimum mean blood pressure (47 mmHg), the average urea (8 mg/dL), and the minimum heart rate (68.31 bpm).

Fig.~\ref{fig:ind_pred_stemi}(c) presented the waterfall plot of a man who died. For this patient, high values of the average urea (32.5 mg/dL), high values of the maximum respiratory rate (37 bpm), and high values of the average creatinine (1.45 umol/L) pushed the model’s output more strongly towards a high probability of mortality ($f(x) = 0.986$). However, Fig.~\ref{fig:ind_pred_stemi}(d) provides detailed predictions of a man who survives. In this case, the feature values that had a greater impact on the model’s output (a low probability, $f(x) = 0.048$) were normal values of the average urea (9 mg/dL), systolic blood pressure (107.84 mmHg), and creatinine (0.7 units/L).

\begin{figure*}[h!]
\centering
\subcaptionbox{Women with a high risk}{\includegraphics[scale=0.24]{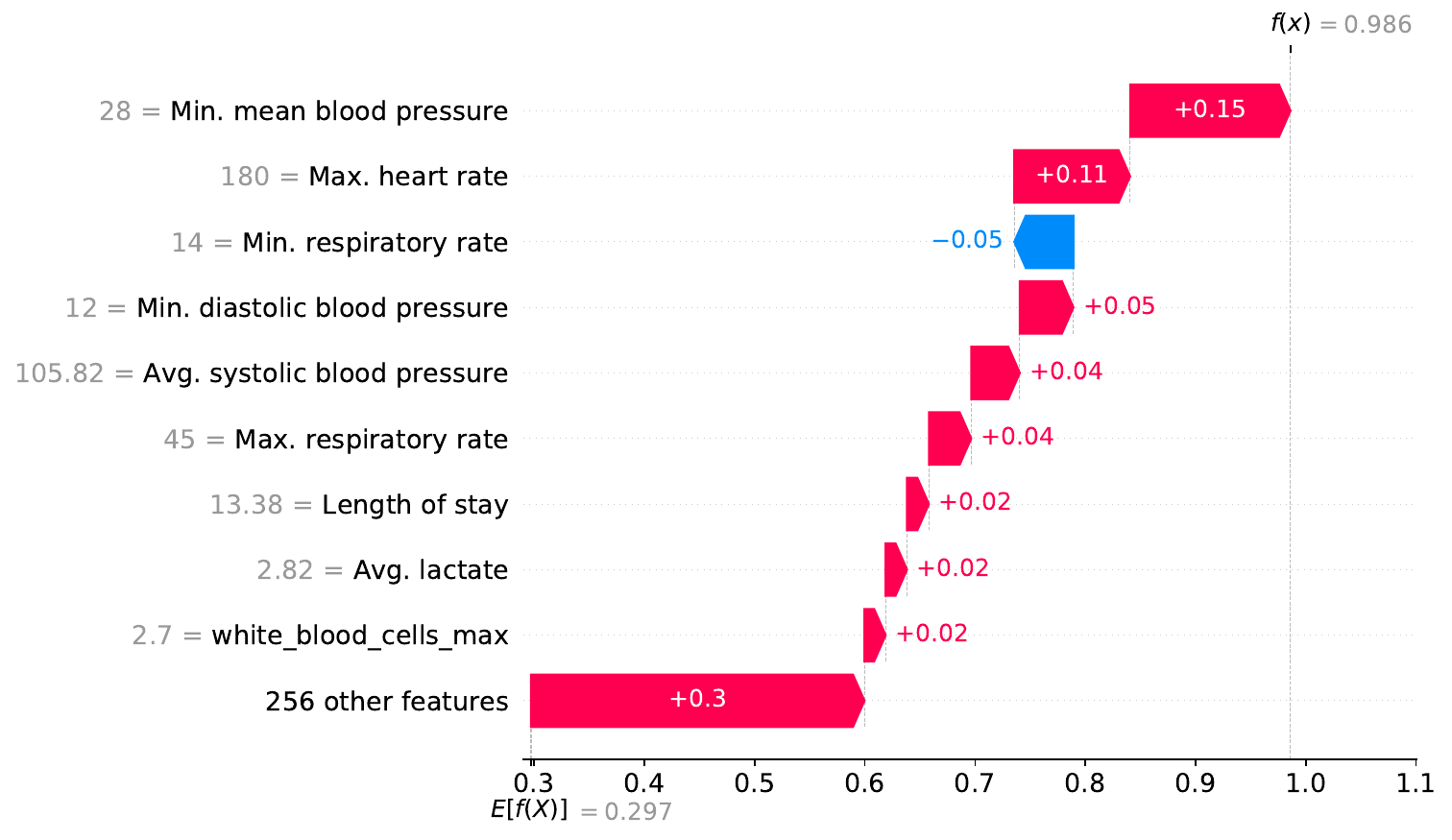}}
\subcaptionbox{Women with a low risk}{\includegraphics[scale=0.24]{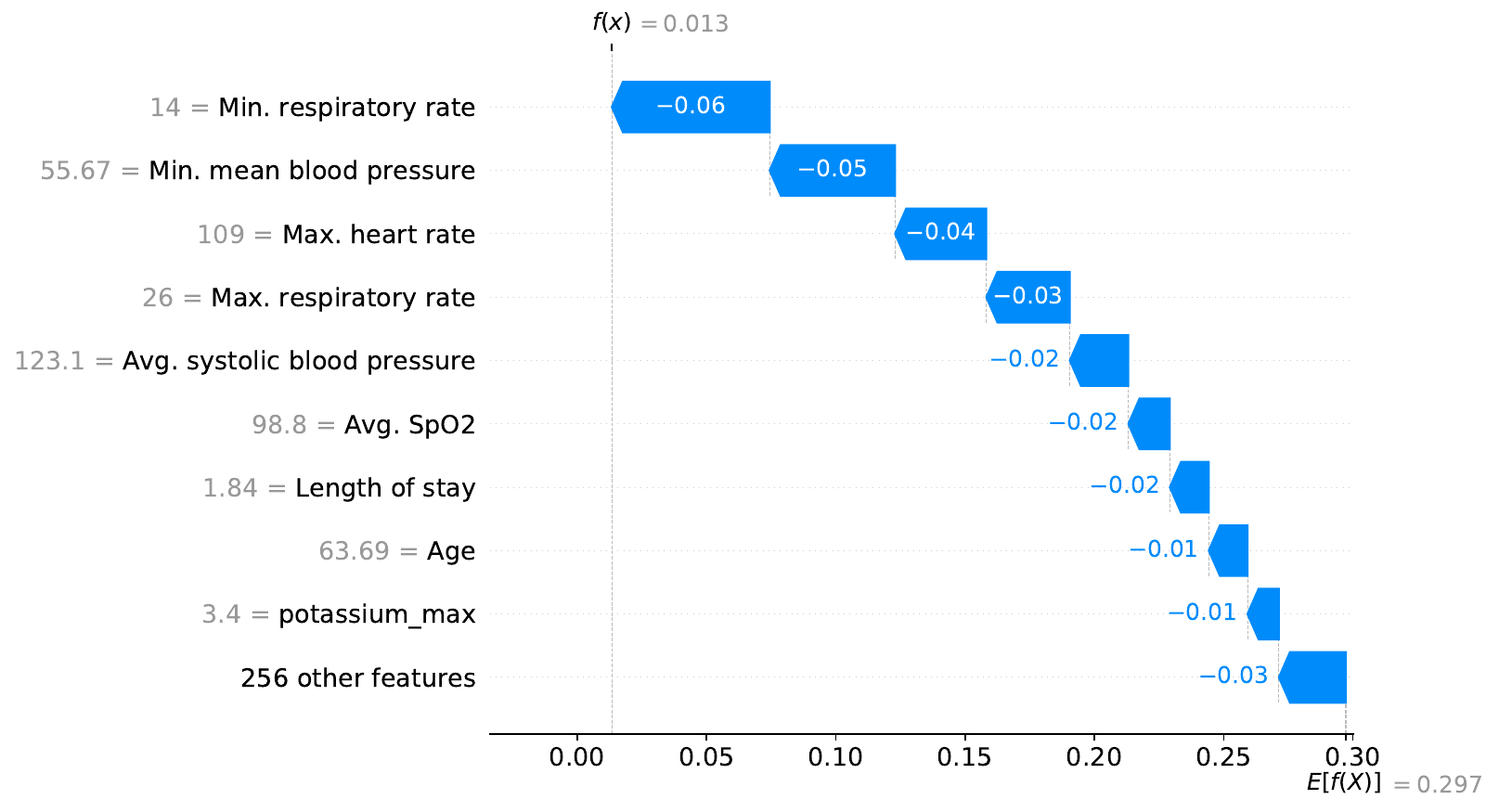}}

\subcaptionbox{Men with a high risk}{\includegraphics[scale=0.24]{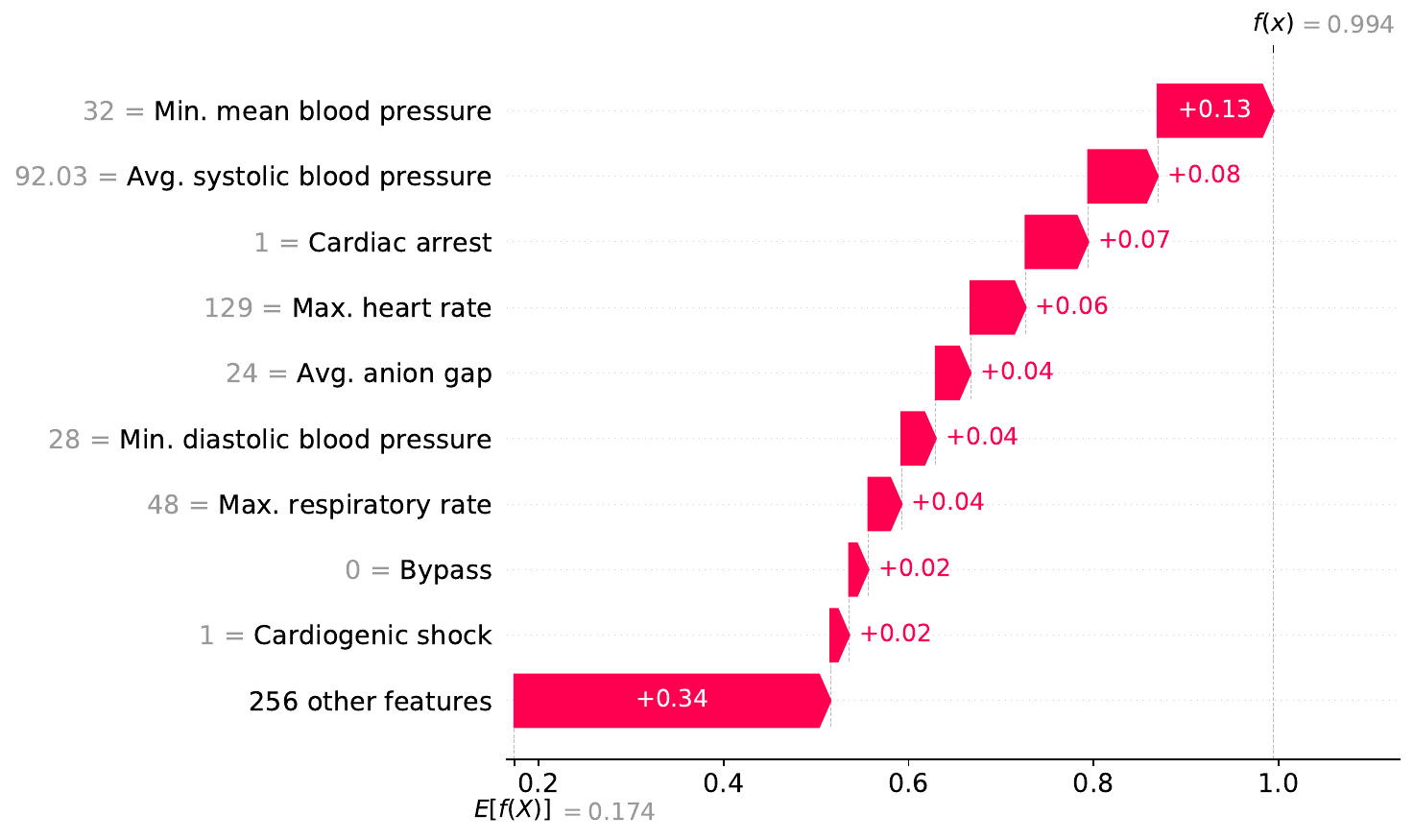}}
\subcaptionbox{Men with a low risk}{\includegraphics[scale=0.24]{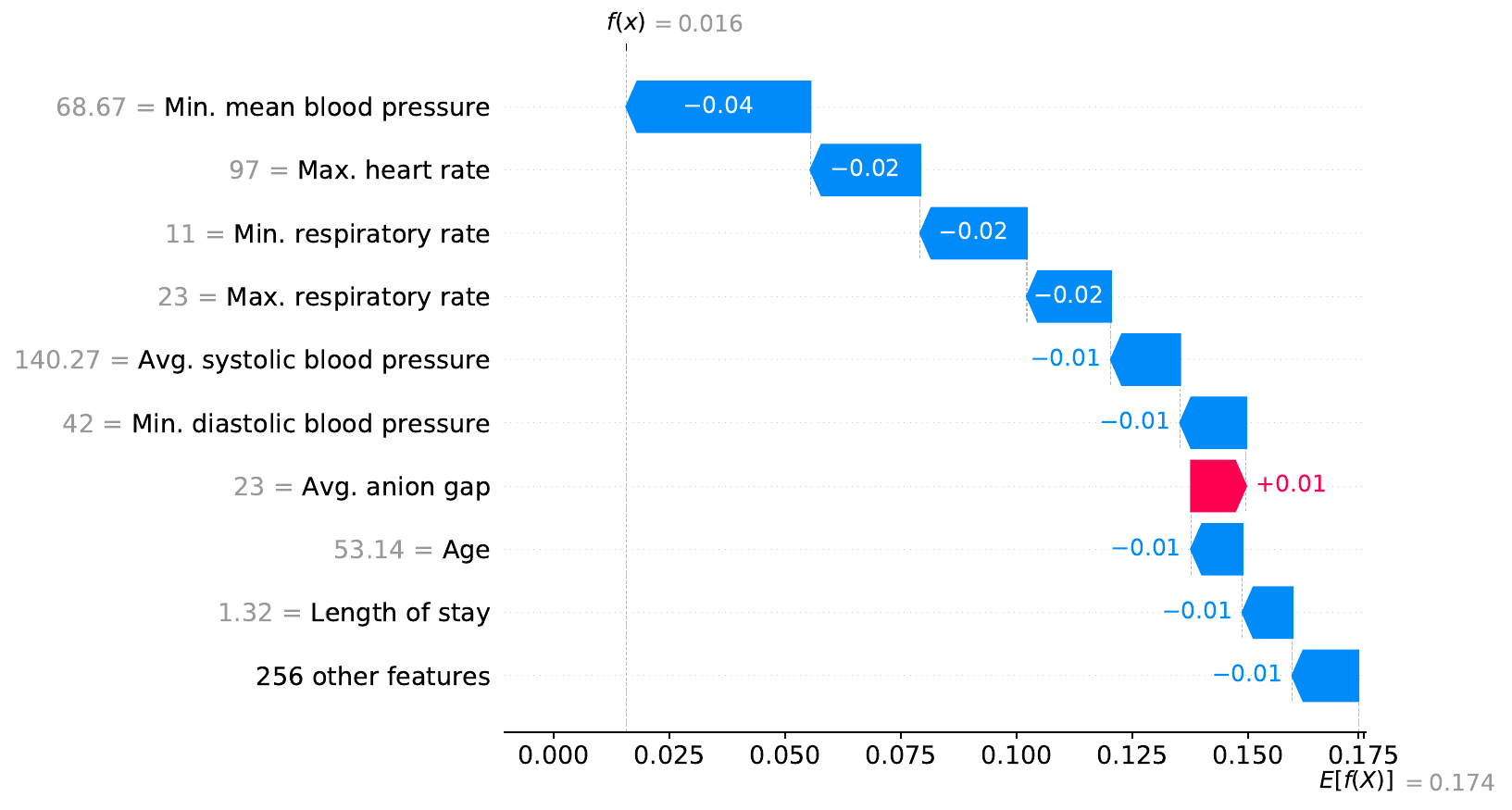}}
\caption{Explanations of individual predictions for patients with NSTEMI.}
\label{fig:ind_pred_nstemi}
\end{figure*}

Fig.~\ref{fig:ind_pred_nstemi} shows the waterfall plots that predictions of four sample patients with NSTEMI. However, Fig.~\ref{fig:ind_pred_nstemi}(a) describes the feature contributions of a woman who died. For this patient, low values of the minimum mean blood pressure (28 mmHg), high values of the maximum heart rate (180 bpm), and low values of the minimum diastolic blood pressure (12 mmHg) were the greatest impact on the high predicted probability ($f(x) = 0.986$). In contrast, Fig.~\ref{fig:ind_pred_nstemi}(b) explained the prediction of a woman who survived. For this patient, the model’s output was a low probability of mortality ($f(x) = 0.013$), where normal values of the minimum respiratory rate (14 bpm), the minimum of blood pressure (55.67 mmHg), and the maximum heart rate (109 bpm) decreased the probability.

Fig.~\ref{fig:ind_pred_nstemi}(c) illustrated the features contributions of a man who died. In this case, the model’s output was a high risk of mortality ($f(x) = 0.994$). The feature values that pushed the prediction towards a higher risk were low values of the minimum mean bloop pressure (32 mmHg), low values of the average systolic blood pressure (92.03 mmHg), and the presence of cardiac arrest. Finally, Fig.~\ref{fig:ind_pred_nstemi}(d) showed the explanation of the prediction for a man who survived. The feature values that had the greatest impact on the low predicted probability ($f(x) = 0.016$) were normal values of the minimum mean blood pressure (68.67 mmHg), normal values of the maximum heart rate, and the minimum respiratory rate (11 bpm).

Thus, for the high-risk patients (Fig.~\ref{fig:ind_pred_stemi}(a) and Fig.~\ref{fig:ind_pred_stemi}(c), and Fig.~\ref{fig:ind_pred_nstemi}(a) and Fig.~\ref{fig:ind_pred_nstemi}(c)), the contribution of the combination of the least important features (bottom row) is significantly higher than the contribution of the top individual features. This suggests that, although some noticeably have higher contributions than others, in some cases, high output values depend on smaller contributions of several features rather than large contributions of a few features.

\subsection{Statistical significance of the identified risk markers}
To assess the significance and coherence of the identified markers using the SHAP approach, we computed their significance in predicting mortality based on a multivariable Cox regression model\footnote{The Python code to fit the Cox model is available at~\url{https://github.com/blancavazquez/Riskmarkers_ACS}}.

We fitted a Cox model with all the STEMI and NSTEMI features and used this model to compute the significance of the top markers identified by the SHAP approach for women, men, and both. Table~\ref{tab:shap_vs_cox} describes the results of these experiments. 

As observed, the Cox model was average creatine kinase MB and the average heart rate were the markers that were statistically significant for both women and men with STEMI. Minimum respiratory rate, minimum lactate, and cardiac arrest were the statistically significant markers only for women. For men only, the average systolic blood pressure, the maximum diastolic blood pressure, and the minimum mean blood pressure were the statistically significant markers. Common statistically significant markers for NSTEMI in both women and men were the average systolic blood pressure, minimum mean blood pressure, and age. Markers that were statistically significant only for women were the maximum diastolic blood pressure, minimum respiratory rate, and cardiac arrest. Only for men were the average heart rate and minimum lactate.

Interestingly, the maximum diastolic blood pressure was a marker that appeared only for men with STEMI, which was also statistically significant in the Cox model. However, for NSTEMI, this marker was statistically significant only for women in the Cox model. Although for both women and men, it ranked among the top of the SHAP feature importance. Overall, the top risk markers found by the SHAP approach (Fig.~\ref{fig:risk_markers} and Fig.~\ref{fig:specific_sex_markers}) were statistically significant in the Cox model.

\begin{table}[h!]
\caption{\label{tab:shap_vs_cox} Statistically significant risk markers for women and men with STEMI and NSTEMI according to the multivariable Cox regression model.}
\begin{center}
\scalebox{0.85}{
\begin{tabular}{lcccc}
\hline
\multicolumn{1}{c}
{\multirow{3}{*}{\begin{tabular}[c]{@{}c@{}}Risk 
\\ marker\end{tabular}}} 
&
\multicolumn{2}{c}{STEMI}
& \multicolumn{2}{c}{NSTEMI}
\\ \cline{2-5} 
\multicolumn{1}{c}{}
& \multicolumn{1}{c}{Women}
& \multicolumn{1}{c}{Men}
& \multicolumn{1}{c}{Women}
& \multicolumn{1}{c}{Men}
\\ \cline{2-5} 
\multicolumn{1}{c}{}
& \multicolumn{1}{c}
{\begin{tabular}[c]{@{}c@{}}p-value\end{tabular}}
& \multicolumn{1}{c}
{\begin{tabular}[c]{@{}c@{}}p-value\end{tabular}}
& \multicolumn{1}{c}
{\begin{tabular}[c]{@{}c@{}}p-value\end{tabular}}
& \multicolumn{1}{c}
{\begin{tabular}[c]{@{}c@{}}p-value\end{tabular}}
\\ \hline
Avg. systolic blood pressure
& 0.02
& \textless{} 0.005*
& \textless{} 0.005*
& \textless{} 0.005*
\\
Avg. creatine kinase MB
& \textless{} 0.005*
& \textless{} 0.005*
& 0.28
& 0.20
\\
Avg. heart rate
& \textless{} 0.005*
& \textless{} 0.005*
& 0.01
& \textless{} 0.005*
\\
Max. diastolic blood pressure
& 0.01
& \textless{} 0.005*
& \textless{} 0.005*
& 0.10
\\
Min. respiratory rate
& \textless{} 0.005*
& 0.50
& \textless{} 0.005*
& 0.05
\\
Min. lactate
& \textless{} 0.005*
& 0.94
& 0.24
&  \textless{} 0.005*
\\
Min. mean blood pressure
& 0.01
& \textless{} 0.005*
& \textless{} 0.005*
& \textless{} 0.005*
\\
Cardiac arrest
& \textless{} 0.005*
& 0.85
& \textless{} 0.005*
& 0.02
\\
Age
& 0.08
& 0.01
& \textless{} 0.005*
& \textless{} 0.005*
\\ \hline
\end{tabular}
}
\end{center}
{
\centering
\footnotesize
* Statistically significant markers (p \textless~0.05).}
\end{table}

\section{Discussion}

The outcomes of the evaluation using ML algorithms for mortality prediction showed that the models trained with all the clinical features achieved a high cross-validated mean AUC. In most settings, XGB outperformed LR, RF, and SVM. Thus, XGB models trained with all the features achieved the highest cross-validated mean AUC. 

Remarkably, models trained with vital signs and laboratory results achieved a high cross-validated mean AUC for STEMI (0.88 and 0.87, respectively) and NSTEMI (0.92 and 0.76, respectively). The results are not surprising as these groups have critical clinical features for both ACS sub-populations that rank very high in the SHAP feature importance computed from the models trained with all the features, e.g., vital signs such as systolic blood pressure, heart rate, respiratory rate, and laboratory results, such as creatine kinase MB, urea, and creatinine.

We identified a set of markers that increased the risk of mortality in women and men using the trained XGB models and the SHAP approach. Moreover, we computed the SHAP feature importance to find the markers that had the highest impact on mortality. We found that vital signs are common risk markers in both women and men with STEMI and NSTEMI. However, some differences are observed in laboratory results, procedures, and complications.

For instance, laboratory results such as creatinine, lactate, anion gap, and creatine kinase mb represent a higher risk for STEMI patients. However, heart bypass (procedure) and cardiac arrest (complication) are a higher impact on mortality in NSTEMI patients. Notably, most of the markers that ranked high in SHAP feature importance were statistically significant according to a multivariable Cox regression model.

SHAP beeswarm and dependence scatter plots show important sex differences in the top risk markers. For example, men with STEMI face a higher risk when they suffer from high levels of urea and low values of creatinine, which could be associated with acute kidney failure (rapid decrease in the renal function manifested by an increase in serum creatinine~\cite{ostermann_acute_2016, singh_acute_2010}).

In contrast, high levels of urea and creatinine have a greater impact on mortality for women with STEMI, which might be associated with chronic kidney failure (persistent damage to the kidneys). Moreover, we found that women with STEMI die younger than men (70 years and 75, respectively), while men with NSTEMI die younger than women (70 and 80 years, respectively). Notably, kidney failure is a known adverse prognostic factor in patients with cardiovascular diseases~\cite{lekston_impaired_2009, shroff_acute_2013}. 

We also distinguish some interesting differences by analyzing the explanations of individual predictions (Fig.~\ref{fig:ind_pred_stemi} and Fig.~\ref{fig:ind_pred_nstemi}). For instance, we found that patients with STEMI often suffered from prolonged thromboplastin times, high values of positive end-expiratory pressure, low values of base excess, and hyponatremia (low levels of sodium). Conversely, NSTEMI patients frequently suffer from hypotension (low values of systolic and diastolic blood pressure), extended lengths of stay, high values of heart rate, and cardiac arrest.

Moreover, two expert cardiologists evaluated the identified markers qualitatively and found them consistent with the clinical routine because they were associated with well-established clinical trends in patients admitted to ICU (e.g., kidney failure, hypotension, and hyponatremia). 

On the other hand, we compared these markers with a longitudinal-cohort study based on the Mexican population, called RENASCA~\cite{borrayo_sanchez_stemi_2018}, which analyzed the risk markers for STEMI and NSTEMI separately. For STEMI, common markers between RENASCA and those identified with the SHAP approach are creatinine, urea, systolic blood pressure, heart rate, creatine kinase mb, acute renal failure, and age. Correspondingly, for NSTEMI, the creatinine, respiratory rate, systolic blood pressure, troponin, and the prevalence of women with advanced age are common between RENASCA and the SHAP approach. Overall, our results show that ML mortality models trained on EHRs are effective in finding significant and coherent sex-specific risk markers of different ACS sub-populations.

\section{Conclusion}
We identified in-hospital mortality markers for women and men in ACS sub-populations from a public database of EHRs using ML models. We trained and validated mortality prediction models and interpreted those with the highest cross-validated mean AUC. We found that ML models trained on EHR data could adequately predict outcomes within the next 24 hours for patients admitted to ICU after suffering a STEMI or NSTEMI. In addition, the identified markers, both general and sex-specific, are relevant and consistent with the clinical routine and with a longitudinal cohort study.

Our findings demonstrate that ML models can discover coherent risk markers, thereby simplifying the identification of clinical markers for different subpopulations. Accordingly, this work could be replicated to extract specific markers in other subpopulations, e.g., age or clinical history, leading to appropriate treatment strategies and better clinical outcomes.

An important limitation of our work is that the EHRs extracted from the MIMIC-III database comprise of patients admitted to different ICUs, and the identified markers could be associated with other conditions that are not directly connected to STEMI and NSTEMI. Therefore, we believe that it would be helpful to restrict the analysis to patients of coronary care units. Another important limitation was that the ML models were trained and evaluated on a rather small population (4,119 patients), which could lead to overfitting. To alleviate this problem, we measured the AUC-ROC and used Repeated Stratified K-Fold Cross-Validation. However, we consider that the models should be trained and evaluated, and the markers should be identified and validated on a larger population to support our findings more strongly.

\section*{Author’s contributions}
Blanca Vazquez conducted research, compiled data, developed the models, and drafted the manuscript. Gibran Fuentes-Pineda directed the research, reviewed the manuscript critically, and directed revisions. Fabian Garcia reviewed the manuscript critically. Gabriela Borrayo and Juan Prohías evaluated all clinical results obtained. All authors read and approved the final manuscript.

\section*{Acknowledgements}
Not applicable.

\section*{Declaration of Competing Interest}
The authors declare that they have no competing interests.

\section*{Funding}
The corresponding author was supported by the National Council for Science and Technology (CONACYT), scholarship number 104442. The funder was not involved in the design, development, or evaluation of the research.

\bibliographystyle{elsarticle-num} 
\bibliography{references}
\end{document}